\documentclass[10pt,twocolumn,letterpaper]{article}
\usepackage{cvpr}    
\usepackage{parskip}
\usepackage[table]{xcolor}
\usepackage{lineno}
\usepackage[accsupp]{axessibility}
\newcommand*{\affmark}[1][*]{\textsuperscript{#1}}

\definecolor{cvprblue}{rgb}{0.0,0.0,1}
\usepackage[pagebackref,breaklinks,colorlinks,allcolors=cvprblue]{hyperref}
\usepackage{multirow}
\definecolor{red}{rgb}{1,0.0,0.0}
\definecolor{violetred}{RGB}{220, 0, 220}

\def\paperID{11539} 
\def\confName{CVPR}
\def\confYear{2025}

\title{ABBSPO: Adaptive Bounding Box Scaling and Symmetric Prior based Orientation Prediction for Detecting Aerial Image Objects}

\author{
Woojin Lee\affmark[1]\footnotemark[1] \quad Hyugjae Chang\affmark[1]\footnotemark[1]\quad Jaeho Moon\affmark[1]\quad Jaehyup Lee\affmark[2]\footnotemark[2] \quad Munchurl Kim\affmark[1]\footnotemark[2]\\
\affmark[1]KAIST \quad
\affmark[2]KNU\\
\small{\{woojin412, hmnc97, jaeho.moon, mkimee\}@kaist.ac.kr} \quad
\small{jaehyuplee@knu.ac.kr}\\
\small{\url{https://kaist-viclab.github.io/ABBSPO_site/}}%
}

\begin{document}

\twocolumn[{
\renewcommand\twocolumn[1][]{#1}%
\maketitle
    \begin{center}
           \includegraphics[width=\linewidth]{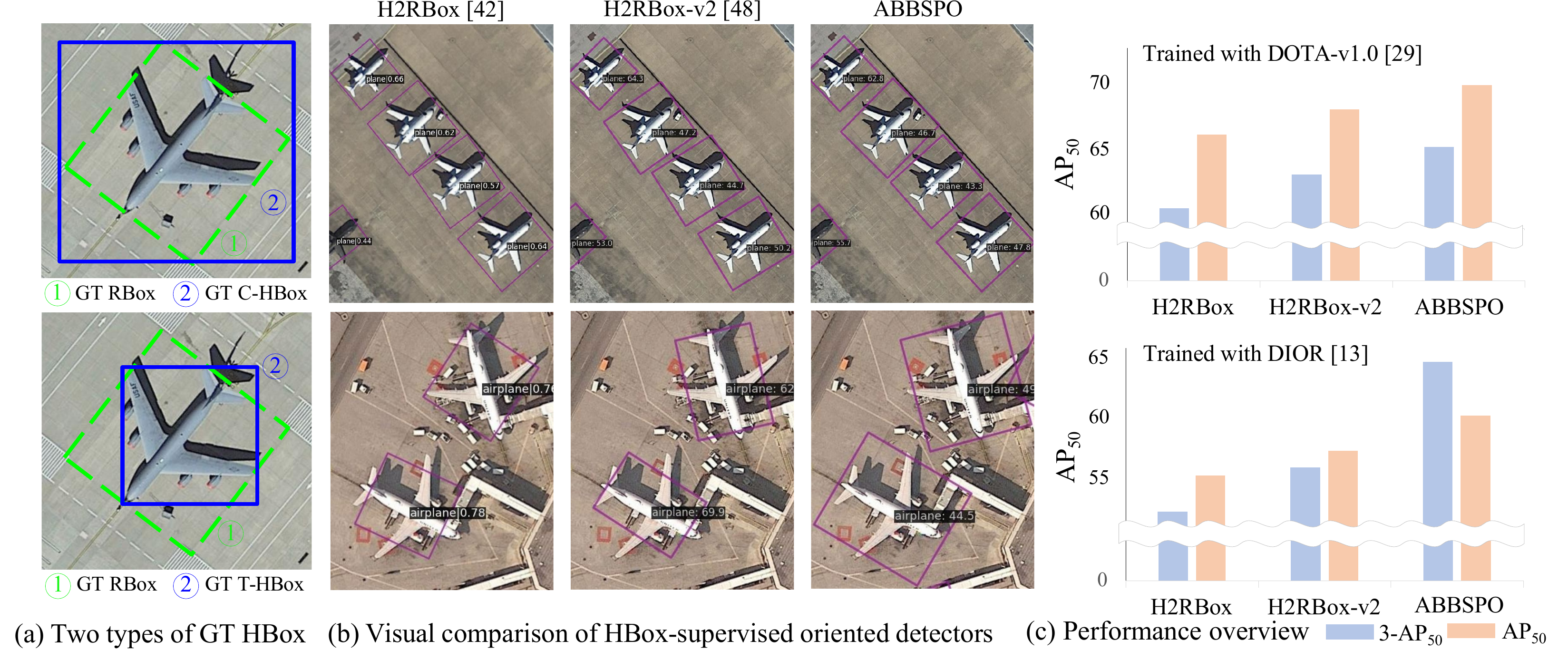}
           \captionof{figure}{
   Performance comparison of HBox-supervised orientated detectors. (a) Top: A coarse horizontal bounding box (C-HBox) (\textcolor{blue}{\textcircled{2}}) and its corresponding rotated bounding box (RBox) (\textcolor{green}{\textcircled{1}}). Bottom: A tight horizontal bounding box (T-HBox)(\textcolor{blue}{\textcircled{2}}) and its corresponding RBox (\textcolor{green}{\textcircled{1}}). (b) Our ABBSPO is capable of accurately detecting both orientations and scales for GT C-HBoxes and T-HBoxes. (c) Average Precision (\(\text{AP}_{50}\)) for H2RBox \cite{h2rbox}, H2RBox-v2 \cite{h2rbox-v2}, and our ABBSPO. 3-\(\text{AP}_{50}\) represents the mean \(\text{AP}_{50}\) for three complex shaped objects: (i) DIOR: `airplane', `expressway service area', and `overpass' and (ii) DOTA: `plane', `swimming pool', and `helicopter'. 
    }
    \label{fig:figure1}
    \end{center}
    \vspace{0.37cm} 
}]

{
  \renewcommand{\thefootnote}
    {\fnsymbol{footnote}}
  \footnotetext[1]{Co-first authors (equal contribution).}
  \footnotetext[2]{Co-corresponding authors.}
}

{
  \renewcommand{\thefootnote}{\arabic{footnote}}  
  \footnotetext[1]{Korea Advanced Institute of Science and Technology (KAIST).}
  \footnotetext[2]{Kyungpook National University (KNU).}
}

\begin{abstract}
Weakly supervised Oriented Object Detection (WS-OOD) has gained attention as a cost-effective alternative to fully supervised methods, providing efficiency and high accuracy. Among weakly supervised approaches, horizontal bounding box (HBox) supervised OOD stands out for its ability to directly leverage existing HBox annotations while achieving the highest accuracy under weak supervision settings. This paper introduces adaptive bounding box scaling and symmetry-prior-based orientation prediction, called ABBSPO that is a framework for WS-OOD. Our ABBSPO addresses the limitations of previous HBox-supervised OOD methods, which compare ground truth (GT) HBoxes directly with predicted RBoxes’ minimum circumscribed rectangles, often leading to inaccuracies. To overcome this, we propose: (i) Adaptive Bounding Box Scaling (ABBS) that appropriately scales the GT HBoxes to optimize for the size of each predicted RBox, ensuring more accurate prediction for RBoxes' scales; and (ii) a Symmetric Prior Angle (SPA) loss that uses the inherent symmetry of aerial objects for self-supervised learning, addressing the issue in previous methods where learning fails if they consistently make incorrect predictions for all three augmented views (original, rotated, and flipped). Extensive experimental results demonstrate that our ABBSPO achieves state-of-the-art results, outperforming existing methods.
\end{abstract}  
\section{Introduction}
\label{sec:intro}
Object detection often leverages supervised learning with ground truth horizontal bounding box labels (GT HBoxes) to locate the objects of interest. However, the usage of GT HBoxes limits the precise localization of the objects with their orientations and tight surrounding boundaries, especially for objects such as airplanes and ships of various orientations in aerial images. To handle object detection as an oriented object detection problem, more precise rotated bounding box labels (GT RBoxes) are required, which is very costly to generate \cite{weak}.
So, to mitigate this challenge, previous methods \cite{weak,wsodet,pointobb,point2rbox,h2rbox,h2rbox-v2} have explored weakly supervised oriented object detection (OOD) that utilizes less expensive forms of annotations, such as image-level, point and HBox annotations. 
Among these, the use of HBoxes is the most popular due to their widespread availability in existing public datasets \cite{fair1m,nwpuvhr-10,xview,rsod,simd,airsarship} to predict the RBoxes for objects of interest. 
So, this approach can detour the costly process of generating GT RBoxes. 

The previous weakly supervised (WS) learning of OOD \cite{h2rbox,h2rbox-v2,Sunetal,eie-det} utilizes GT HBoxes in the forms of \textit{coarse} HBoxes, called GT C-HBoxes, as supervision to compare with the HBoxes derived as the minimum circumscribed rectangles from the predicted RBoxes by their OOD models.  As shown in the upper figure of Fig.~\ref{sec:intro}-(a), the GT C-HBoxes are defined as \textit{coarse} horizontal bounding boxes that loosely encompass the boundaries of objects (not tightly bounded). The GT HBoxes of the DOTA \cite{dota} dataset are in the forms of C-HBoxes which are derived as the minimum circumscribed horizontal bounding boxes of their GT RBoxes.
However, when the previous OOD methods \cite{h2rbox, h2rbox-v2} are supervised with the other  GT HBoxes that are in the form of \textit{tight} HBoxes, called GT T-HBoxes (e.g. DIOR dataset \cite{dior}), as shown in the bottom figure of Fig. \ref{sec:intro}-(a), we found that their performances are significantly degraded because GT T-HBoxes tend to have different scales, compared to those of GT C-HBoxes (see Fig.~\ref{sec:intro}-(c)). As shown in Fig.~\ref{sec:intro}-(b), this causes the previous methods to predict either RBoxes with accurate orientations but inaccurate scales smaller than the sizes of their corresponding objects, or the RBoxes with inaccurate (close to horizontal) orientations but somewhat accurate scales (almost the same as HBoxes).

To overcome the above limitations of the previous WS-OOD methods, we propose an adaptive bounding box scaling and symmetry-prior-based orientation prediction, called as ABBSPO, as a WS-OOD framework that can be effectively trained with either GT C-HBoxes or GT T-Hboxes for aerial images. 
For this, (i) a novel Adaptive Bounding Box Scaling (ABBS) module is designed to have the flexibility of adjusting the GT HBoxes for each object into random sizes and then selecting the optimal scaled GT HBoxes that allow it to encompass the predicted RBoxes. 
Note that the previous methods are not possible to have such flexibility for the adjustment of GT HBoxes; (ii) An angle learning module is proposed in a self-supervised manner that utilizes the symmetric priors of the objects that open appear in top-down views of aerial images. 
As shown in Figs~\ref{sec:intro}-(b) and (c), Our proposed method predicts accurate orientation and surrounding boxes of objects for both cases of using GT C-HBoxes and GT T-HBoxes, outperforming the previous methods in angle accuracy and localization in terms of average precision (AP).
Our contributions are summarized as:        

\begin{itemize}[leftmargin=1.25em]
    \item To the best of our knowledge, our work is the \textit{first} to address the limitations of previous weakly supervised OOD learning methods with T-HBoxes as GT. To overcome this, we propose a novel weakly supervised OOD method that can be effectively trained with T-HBoxes or C-HBoxes that can be cheaply annotated as GT;  
    
    \item The adaptive bounding box scaling (ABBS) module is proposed to flexibly adjust the HBox (GT) for each object toward an appropriately scaled HBox. This allows part of the predicted RBoxes to place outside the T-HBox (GT), yielding precise RBox prediction;

    \item A symmetric prior angle (SPA) loss is presented to enhance the orientation prediction accuracy by leveraging the symmetric priors of the objects in aerial images;

    \item Our method significantly outperforms the state-of-the-art OOD methods using weakly supervised learning with HBoxes (GT) for aerial datasets.
\end{itemize}

\section{Related Work}
\label{sec:related_works}

\begin{figure*} [htbp]
    \centering
    \includegraphics[width=\linewidth]{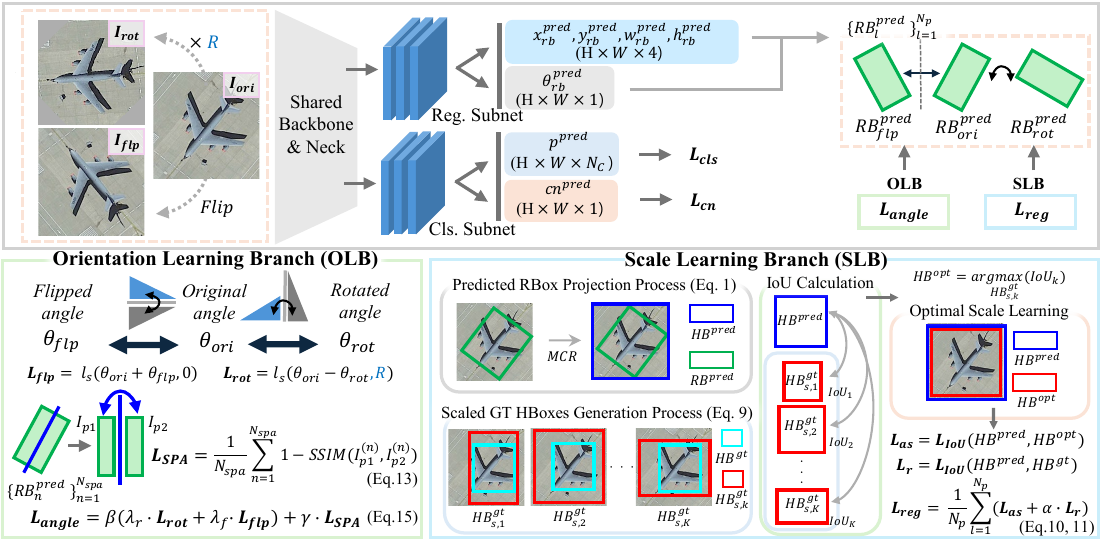}
    \vspace{-0.5cm}
    \caption{\textbf{Overall pipeline of our ABBSPO framework.} 
    Our ABBSPO leverages weakly supervised learning from HBox annotations to accurately predict RBoxes. The framework incorporates the Orientation Learning Branch (OLB) for precise angle estimation, using the Symmetric Prior Angle (SPA) loss, 
    and the Scale Learning Branch (SLB) for optimal scale adjustment via the Adaptive Bounding Box Scaling (ABBS) module. The framework supports both C-HBox and T-HBox ground truths, ensuring robust and accurate predictions.}
    \vspace{-0.3cm}
    \label{fig:figure2}
\end{figure*}

\subsection{RBox-supervised Oriented Object Detection}
\label{subsec:RBox_supervised_aod}

Oriented Object Detection (OOD) has gained significant attention, leading to extensive research in RBox-supervised methods (using GT RBoxes) such as Rotated RetinaNet \cite{retinanet}, Rotated FCOS \cite{fcos}, \(\text{R}^3\)Det \cite{r3det}, ROI Transformer \cite{RoITransformer}, ReDet \cite{redet}, and \(\text{S}^2\)A-Net \cite{s2a-net}. Rotated FCOS \cite{fcos} improves OOD performance by introducing center-ness, which assigns weights to samples based on their proposal locations, thereby emphasizing well-positioned proposals. Oriented-RepPoints methods \cite{reppoints,reppoints-v2,orientedreppoints}, in contrast, utilize flexible receptive fields to extract key object points. However, a common challenge in RBox-supervised OOD methods is the boundary discontinuity problem that arises from the definition and prediction of angle parameters ($\theta$) \cite{arbitrary,rethinking}. To address this, several methods modified the ways of defining the RBox representations, such as Gaussian distributions \cite{csl,dcl,mgar,gf-csl,gwd,kld,gaussiandistributions,kfiou}, thereby avoiding straightforward regression of angle parameters. On the other hand, in weakly supervised learning, the boundary discontinuity issue does not arise thanks to the absence of direct RBox supervision, allowing for more stable angle predictions without the need for complex mitigation strategies.

\subsection{Weakly-supervised Orientd Object Detection}
\label{subsec:weakly_supervised_aod}
Weakly supervised OOD methods learn to predict RBoxes without directly utilizing GT RBoxes. 
The approaches in this domain are primarily categorized based on the types of labels they employ: image-based \cite{wsodet}, point-based \cite{pointobb,point2rbox}, and HBox-based \cite{h2rbox,h2rbox-v2}.

\noindent\textbf{Image-based supervision.} 
WSODet \cite{wsodet}, aims to generate pseudo-RBoxes without explicit localization supervision, thus encountering significant limitations when relying solely on image labels, especially for the scenes with numerous and diverse object types.

\noindent\textbf{Point-based supervision.} 
By leveraging one representative point at the center location as a label for each object, point label-based methods offer the advantage of being cost-effective \cite{points1,point2,point3,point4,point5,point6}. PointOBB \cite{pointobb} estimates angles from geometric relationships across original, rotated, and flipped views, and determines scales by analyzing proposal distributions between original and scaled input images. 
PointOBB-v2 \cite{pointobbv2} improves single point supervision by refining pseudo-label generation, leading to enhanced efficiency and accuracy.
Point2RBox \cite{point2rbox} employed fundamental patterns as priors to guide the regression of RBoxes. 
Point-based methods are cost-effective and straightforward, but still struggle with limited supervision.

\noindent\textbf{HBox-based supervision.} 
As annotating HBoxes is more straightforward than RBoxes, the HBox-supervised OOD has gained increasing attention in recent studies.
H2RBox \cite{h2rbox} utilized rotated views from the original view and provided self-supervision of object orientations without requiring the GT angles. 
H2RBox-v2 \cite{h2rbox-v2} expanded the use of geometric relationships between views by adding flipped views. 
These methods learn to predict RBoxes by converting minimum circumscribed HBoxes that encompass the predicted RBoxes to directly compare IoU with the GT HBoxes, thereby enabling HBox-supervised OOD. 
However, these methods only guarantee performance when trained with GT C-HBoxes that are derived from GT RBoxes.
These methods struggle to learn precise OOD when being trained with GT T-HBoxes, because of the significant gap between the GT T-HBoxes and the HBoxes derived from predicted RBoxes. 
To address the issue, we propose an ABBS module to effectively handle both types of GTs (C-HBoxes and T-HBoxes).

\section{Method}
\subsection{Overall Pipeline}
Weakly supervised OOD aims to predict RBoxes using less expensive annotations such as HBoxes. Existing methods such as H2RBox \cite{h2rbox} and its improved version H2RBox-v2 \cite{h2rbox-v2} have laid the foundation for directly predicting RBoxes from HBoxes. Our proposed pipeline builds upon the H2RBox-v2 \cite{h2rbox-v2} framework to effectively enable weakly supervised OOD from either C-HBoxes or T-Boxes.
Figure~\ref{fig:figure2} depicts the conceptual framework of our weakly supervised OOD method. 
Given an input image $\text{I}_\textit{ori}$ and its rotated and flipped version $\text{I}_\textit{rot}$ and $\text{I}_\textit{flp}$, our proposed pipeline obtains the RBox for each input view ($\text{I}_\textit{ori}$, $\text{I}_\textit{rot}$, $\text{I}_\textit{flp}$), including the center position ($\textit{x,y}$), size ($\textit{w,h}$), angle ($\theta$), class scores ($\textit{p}$), and the center-ness ($\textit{cn}$).
To classify each detected object, we follow FCOS~\cite{fcos} by supervising both the classification (\(\textit{p}\)) and the center-ness (\(\textit{cn}\)). The angle ($\theta$) prediction is obtained by using the method proposed in PSC~\cite{psc}. Our contribution mainly lies in the supervision for localization, consisting of two branches: a scale learning branch (SLB) and an orientation learning branch (OLB). 

In the SLB, the adaptive bounding box scaling (ABBS) module addresses the relationships between GT HBoxes and predicted RBoxes.
This ABBS module provides proper minimum circumscribed rectangle for the accurately predicted RBoxes, by adaptively scaling the HBoxes based on a predefined scale range. 
The OLB guides accurate prediction of object orientation by utilizing three input views ($\text{I}_\textit{ori}$, $\text{I}_\textit{rot}$, $\text{I}_\textit{flp}$), following H2RBox-v2 \cite{h2rbox-v2}.
Additionally, the OLB utilizes these orientation predictions for our symmetric prior angle (SPA) loss, which leverages the inherent left-right symmetry of objects in aerial images. 
The SPA loss enforces to further adjust the orientations of the predicted RBoxes to be aligned with the orientations of the symmetric objects such as airplanes, ships, ground track fields etc. 

\subsection{Adaptive Bounding Box Scaling Module}  
\label{subsec:adaptive_bounding_box_scaling_module}

In Fig.~\ref{fig:figure2}, `Scale Learning Branch' illustrates the conceptual process of our adaptive bounding box scaling module (ABBS) module. In HBox-supervised OOD learning, the predicted RBoxes (\(\textit{RB}^{\text{pred}}\)) must be compared with the GT HBoxes (\(\textit{HB}^{\text{gt}}\)). Since they cannot be directly compared, \(\textit{RB}^{\text{pred}}\) is first converted to \(\textit{HB}^{\text{pred}}\), defined as the minimum circumscribed HBox of \(\textit{RB}^{\text{pred}}\) as:
\vspace{-0.1cm}
\begin{equation}
    \textit{HB}^{\text{pred}} = \textit{MCR}(\textit{RB}^{\text{pred}}),
\vspace{-0.1cm}
\end{equation}
where \(\textit{MCR}(\cdot)\) is an operator converting \(\textit{RB}\) to the minimum circumscribed \(\textit{HB}\), allowing \(\textit{HB}^{\text{pred}}\) to be compared with \(\textit{HB}^{\text{gt}}\). 
\(\textit{HB}^{\text{pred}}\) and \(\textit{RB}^{\text{pred}}\) are given by:
\begin{equation}
\vspace{-0.1cm}
    \begin{split}
    &\textit{RB}^{\text{pred}} = [x_{\textit{rb}}^{\text{pred}}, y_{\textit{rb}}^{\text{pred}}, w_{\textit{rb}}^{\text{pred}}, h_{\textit{rb}}^{\text{pred}}, \theta_{\textit{rb}}^{\text{pred}}], \\
    &\textit{HB}^{\text{pred}} = [x_{\textit{hb}}^{\text{pred}}, y_{\textit{hb}}^{\text{pred}}, w_{\textit{hb}}^{\text{pred}}, h_{\textit{hb}}^{\text{pred}}],
    \end{split}
\vspace{-0.1cm}
\end{equation}
where \((x_{\textit{rb}}^{\text{pred}}, y_{\textit{rb}}^{\text{pred}})\) and \((x_{\textit{hb}}^{\text{pred}}, y_{\textit{hb}}^{\text{pred}})\) are the centers of \(\textit{RB}^{\text{pred}}\) and \(\textit{HB}^{\text{pred}}\), respectively. The width \(\textit{w}\) and height \(\textit{h}\) of \(\textit{HB}^{\text{pred}}\) can be computed as:
\begin{equation}
    \begin{split}
            & \textit{w}_{\textit{hb}}^{\text{pred}} = \textit{w}_{\textit{rb}}^{\text{pred}} |\cos \theta_{\textit{rb}}^{\text{pred}}| + \textit{h}_{\textit{rb}}^{\text{pred}} |\sin \theta_{\textit{rb}}^{\text{pred}}|, \\
            & \textit{h}_{\textit{hb}}^{\text{pred}} = \textit{w}_{\textit{rb}}^{\text{pred}} |\sin \theta_{\textit{rb}}^{\text{pred}}| + \textit{h}_{\textit{rb}}^{\text{pred}} |\cos \theta_{\textit{rb}}^{\text{pred}}|.
    \end{split}
\end{equation}
If \(\textit{RB}^{\text{opt}}\) is defined as the tightly surrounding object boundary RBox with the precise orientation, then we have \(\textit{HB}^{\text{opt}} = \textit{MCR}(\textit{RB}^{\text{opt}})\) for the `Predicted RBox Projection Process' block in the SLB of Fig.~\ref{fig:figure2}. When the size of \(\textit{HB}^{\text{gt}}\) is larger or smaller than that of \(\textit{HB}^{\text{opt}}\), the model needs to adaptively adjust and find the optimal scale within a predefined range of scale variations.
We propose an ABBS module that estimates \(\textit{RB}^{\text{opt}}\) by adaptively adjusting the scale of \(\textit{HB}^{\text{gt}}\). Notably, even if \(\textit{RB}^{\text{pred}}\) is accurately estimated, its \(\textit{HB}^{\text{pred}}\) may not overlap well with \(\textit{HB}^{\text{gt}}\), leading to a low Intersection over Union (IoU) value. Enforcing \(\textit{HB}^{\text{pred}}\) to match \(\textit{HB}^{\text{gt}}\) can cause a misalignment with \(\textit{RB}^{\text{pred}}\)  because \(\textit{HB}^{\text{gt}}\) may not be ideal in estimating \(\textit{RB}^{\text{opt}}\). To address this, our ABBS module adaptively scales and adjusts \(\textit{HB}^{\text{gt}}\) in the context of \(\textit{RB}^{\text{pred}}\), rather than forcing \(\textit{HB}^{\text{pred}}\) to match \(\textit{HB}^{\text{gt}}\).

\begin{figure}
    \centering
    \includegraphics[width=0.8\linewidth]{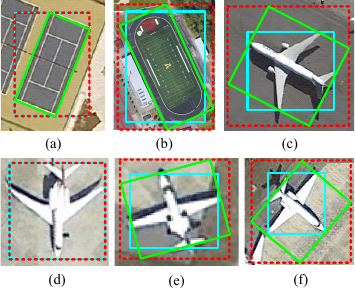}
    \vspace{-0.3cm}
    \caption{Analysis of scale adjustment function (\(\textit{f}(\cdot)\)) based on the shape and angle of objects. (a) rectangular shape, (b) rounded rectangular shape, (c) complex shape, (d) horizontal orientation, (e) slightly tilted orientation, (f) diagonal orientation. The cyan solid box,  green solid box and red dotted box represent GT T-HBox, RBox and adjusted GT HBox, respectively.}
    \vspace{-3mm}
    \label{fig:figure3}
\end{figure}

For the detailed explanation of our ABBS module, we first define a set of scaled versions of \(\textit{HB}^{\text{gt}}\) for the `Scaled GT HBoxes Generation Process' in the SLB of Fig.~\ref{fig:figure2} as:
\vspace{-0.1cm}
\begin{equation} \label{eq:4}
    \begin{split}
        \textbf{HB}^{\text{gt}}_{\text{s}} = \{\textit{HB}^{\text{gt}}_{\text{s,1}}, \textit{HB}^{\text{gt}}_{\text{s,2}}, 
        \cdots,
        \textit{HB}^{\text{gt}}_{\text{s,K}}\},
    \end{split}
\end{equation}
\vspace{-0.1cm}
where \(\textit{HB}^{\text{gt}}_{\text{s},k}\) is the k-th scaled version of \(\textit{HB}^{\text{gt}}\) and K is the total number of scaled variations of \(\textit{HB}^{\text{gt}}\). \(\textbf{HB}^{\text{gt}}_{\text{S}}\) is determined as the combinations of \textit{angle-adjusted} width and height scale factors, \(\{\textit{s}^{\textit{ w}}_{\textit{adj,i}}\}^{\textit{N}_s}_{\textit{i}=1}\) and  \(\{\textit{s}^{\textit{ h}}_{\textit{adj,j}}\}^{\textit{N}_s}_{\textit{j}=1}\), that are transformed from basic width and height scale factors,\(\{\textit{s}^{\textit{ w}}_{\textit{i}}\}^{\textit{N}_s}_{\textit{i}=1}\) and  \(\{\textit{s}^{\textit{ h}}_{\textit{j}}\}^{\textit{N}_s}_{\textit{j}=1}\) by considering angle prediction. Basic scale factors are uniformly spaced in a predefined scale range as:     
\vspace{-0.5cm}
\begin{equation}\label{eq:5}
    \begin{split}
        & \textit{S}_{\textit{w}} = \{\textit{s}^{\textit{ w}}_{\text{1}}, \textit{s}^{\textit{ w}}_{\text{2}}, \cdots, \textit{s}^{\textit{ w}}_{\textit{N}_s}\}, \text{ }\textit{S}_{\textit{h}} = \{\textit{s}^{\textit{ h}}_{\text{1}}, \textit{s}^{\textit{ h}}_{\text{2}}, \cdots, \textit{s}^{\textit{ h}}_{\textit{N}_s}\},
    \end{split}
\end{equation}
\vspace{-0.1cm}
where \(\textit{S}_\textit{w}\) and \(\textit{S}_\textit{h}\) are the sets of basic width and height scale factors, respectively. 
\(\textit{s}^{\textit{ w}}_{\textit{i}}\) and \(\textit{s}^{\textit{ h}}_{\textit{i}}\) are calculated as:
\vspace{-0.1cm}
\begin{equation}\label{eq:6}
    \begin{split}
        &{\textit{s}^{\textit{ w}}_{\textit{i}}} = {\textit{s}^{\textit{ w}}_{1}} + ({\textit{s}^{\textit{ w}}_{\textit{N}_s}}-{\textit{s}^{\textit{ w}}_{1}})/({\textit{N}_s}-1)\cdot(\textit{i}-1),\\
        &{\textit{s}^{\textit{ h}}_{\textit{j}}} = {\textit{s}^{\textit{ h}}_{1}} + ({\textit{s}^{\textit{ h}}_{\textit{N}_s}}-{\textit{s}^{\textit{ h}}_{1}})/({\textit{N}_s}-1)\cdot(\textit{j}-1),
    \end{split}
\end{equation}
where \({\textit{s}^{\textit{w}}_{\textit{N}_s}} = {\textit{s}^{\textit{h}}_{\textit{N}_s}}\) is the predefined largest basic scale factor for both width and height of \(\textit{HB}^{\text{gt}}\), and \({\textit{N}_s}\) is the number of uniform quantization for the both range [\({\textit{s}^{\textit{w}}_{1 }}\text{..}{\textit{s}^{\textit{w}}_{\textit{N}_s}}\)] and [\({\textit{s}^{\textit{h}}_{1 }}\text{..}{\textit{s}^{\textit{h}}_{\textit{N}_s}}\)].
In order to generate \(\textit{HB}^{\text{gt}}_{\textit{s,i}}\), we transform the basic width and height scale factors, \(\{\textit{s}^{\textit{w}}_{\textit{i}}\}^{\textit{N}_s}_{\textit{i}=1}\) and \(\{\textit{s}^{\textit{h}}_{\textit{j}}\}^{\textit{N}_s}_{\textit{j}=1}\), into \textit{angle-adjusted} width and height scale factors, \(\{\textit{s}^{\textit{w}}_{\textit{adj,i}}\}^{\textit{N}_s}_{\textit{i}=1}\) and \(\{\textit{s}^{\textit{h}}_{\textit{adj,j}}\}^{\textit{N}_s}_{\textit{j}=1}\), using the predicted angle \(\theta^{\text{pred}}\) through the scale adjustment function \( f \):
\vspace{-0.2cm}
\begin{equation}
    \label{eq:7}
    \begin{split}
        \textit{s}^{\textit{ w}}_{\textit{adj,i}}\ = \textit{f}(\theta_{\textit{rb}}^{\text{pred}},  {\textit{s}^{\textit{ w}}_{\textit{i}}}),   
         \textit{ s}^{\textit{ h}}_{\textit{adj,j}}\ = \textit{f}(\theta_{\textit{rb}}^{\text{pred}}, {\textit{s}^{\textit{ h}}_{\textit{j}}}).
    \end{split}
\end{equation}
\vspace{-0.1cm}
To define \( f(\cdot) \), it’s essential to consider the object types and rotation angles. Fig.~\ref{fig:figure3} shows the effect of scale adjustments on T-HBoxes for three object types: (i) For rectangular objects like tennis courts (Fig.~\ref{fig:figure3}-(a)), the adjusted T-HBox (red dotted box) aligns precisely with the GT T-HBox (cyan solid box) and tightly circumscribes the optimal RBox (green solid box); (ii) For rounded rectangular objects (Fig.~\ref{fig:figure3}-(b)), the optimal RBox slightly exceeds the GT T-HBox; (iii) Complex shapes like airplanes (Fig.~\ref{fig:figure3}-(c)) show a larger discrepancy, with parts of the optimal RBox lying outside the GT T-HBox. Furthermore, scale adjustments also depend on rotation angles: (i) Fig.~\ref{fig:figure3}-(d) for a vertically (or horizontally) aligned airplane, the GT T-HBox and optimal RBox are identical; (ii) For Fig.~\ref{fig:figure3}-(e) with a small rotation angle, they differ slightly; (iii) For Fig.~\ref{fig:figure3}-(f) with a larger angle, the difference is more pronounced. Therefore, to take the object's shape types and orientation degrees into account for the scale adjustment for the widths and heights of T-HBoxes, \( f(\cdot) \) in Eq.~\ref{eq:7} is defined as:   
\vspace{-0.1cm}
\begin{equation}\label{eq:8}
    \begin{split}
        \textit{f}(\theta,  {\textit{s}}) =
        \begin{cases}
            \frac{4}{\pi}({\textit{s}}-1)\cdot{\theta}+1, &\text{if } 0 \leq {\theta} < \frac{\pi}{4}, \\
            \frac{4}{\pi}(1-{\textit{s}})\cdot{\theta}+ (2s - 1), &\text{if } \frac{\pi}{4} \leq {\theta} < \frac{\pi}{2},
        \end{cases}
    \end{split}
\end{equation}
where the angle range is set to \(\theta \in [0, \pi/2)\) due to the periodicity of the angle. According to \( f(\cdot) \) and Eq.~\ref{eq:7}, \(\textit{HB}^{\text{gt}}_{\textit{s,k}}\) in \(\textbf{HB}^{\text{gt}}_{\textit{s}}\) can be expressed as:
\begin{equation}
\vspace{-0.2cm}
    \label{eq:9}
    \begin{split}
        \textit{HB}^{\text{gt}}_{\textit{s,k}} = [\textit{x}^{\text{gt}}, \textit{y}^{\text{gt}}, \textit{w}^{\text{gt}} \cdot \textit{s}^{\textit{ w}}_{\textit{adj,i}}, 
        \textit{ h}^{\text{gt}} \cdot \textit{s}^{\textit{ h}}_{\textit{adj,j}}],
    \end{split}
\vspace{-0.1cm}
\end{equation}
where (\(\textit{x}^{\text{gt}}, \textit{y}^{\text{gt}}\)) is the center point, and \(\textit{w}^{\text{gt}}\) and \(\textit{h}^{\text{gt}}\) are width and height of \(\textit{HB}^\text{gt}\).
As shown in the `IoU Calculation' and `Optimal Scale Learning' blocks in the SLB of Fig.~\ref{fig:figure2}, \(\textit{HB}^{\text{opt}}\) among \(\{\textit{HB}^{\text{gt}}_{\textit{s,k}}\}^{\textit{K}}_{\textit{k}=1}\) can be determined which minimizes the IoU loss for all proposals by an ABBS loss as:
\begin{equation}
\vspace{-0.2cm}
\label{eq:abbs_loss}
\mathcal{L}_{\text{as}} = \frac{1}{N_p}\sum_{l=1}^{N_p} \min_{\substack{s^w_i \in S_w, \\ s^h_j \in S_h}}\mathcal{L}_{\text{IoU}} \Big( \textit{HB}^{\text{pred}}_{\textit{l}}, \textit{HB}^{\text{gt},\textit{l}}_{\textit{s,k}}(s^w_i, s^h_j) \Big),
\vspace{-0.2cm}
\end{equation}
where \(\textit{N}_p\) is the total number of proposals for input \textbf{I}. \(\textit{HB}^{\text{pred}}_{\textit{l}}\) is \(\textit{HB}^{\text{pred}}\) for \(\textit{l}\)-th proposal, and \(\textit{HB}^{\text{gt},\textit{l}}_{\textit{s,k}}(\textit{s}^{\textit{ w}}_{\textit{i}}, \textit{s}^{\textit{ h}}_{\textit{j}})\) is  \(\textit{k}\)-th scaled \(\textit{HB}^{\text{gt}}\), as \(\textit{HB}^{\text{gt}}_{\textit{s,k}}\), whose width and height are scaled for \(\textit{l}\)-th proposal according to Eq.~\ref{eq:7} to Eq.~\ref{eq:9}. Finally, by adding a regularization term using the IoU loss between \(\textit{HB}^{\text{pred}}\) and non-scaled \(\textit{HB}^{\text{gt}}\), we formed the regression loss as: 
\vspace{-0.05cm}
\begin{equation}
\label{eq:regression_loss}
    \mathcal{L}_{\text{reg}} = \mathcal{L}_{\text{as}} +\alpha\cdot({1}/{N_p})\textstyle\sum^{N_p}_{l=1}\mathcal{L}_{\text{IoU}} \Big( \textit{HB}^{\text{pred}}_{\textit{l}}, \textit{HB}^{\text{gt},\textit{l}} \Big),
\end{equation}
where \(\alpha\) is a hyperparameter which is set to 0.01 by default.
\begin{figure}[t]
    \centering
    \includegraphics[width=\linewidth]{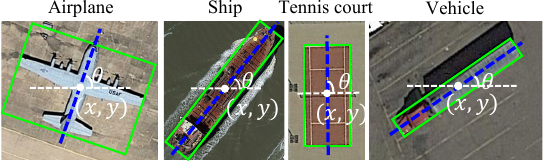}
    \vspace{-0.5cm}
    \caption{\textbf{Examples of symmetric objects in aerial images.} In SPA loss, the \textit{x},\textit{y} coordinates and angle $\theta$ are used to define the symmetry axis, splitting the object into two parts for comparison.}
    \vspace{-0.3cm}
    \label{fig:figure4}
\end{figure}

\subsection{Symmetric Prior Angle Loss}
\label{subsec:symmetric_angle_refinement_module}
In aerial images, objects such as airplanes, ships, tennis courts, and vehicles are often captured from top-down viewpoints, where most of these objects exhibit symmetries in their appearance, as shown in Fig.~\ref{fig:figure4}. 

In the previous pipelines \cite{h2rbox-v2,h2rbox}, both the regression loss associated with the bounding box's center point, width, and height, and the angle loss for accurate angle prediction were trained in a balanced manner. 
However, they tended to inaccurately predict the angles by maximizing the bounding box's IoUs at the same time. 
This issue stems from the fact that, since the angles could not be directly supervised due to the absence of angle annotations, the angles were indirectly supervised from augmented views with rotations and flips. 
This is problematic because, when the difference between two predicted angles for the same object in the original view and its rotated view are equal to the rotation angle applied for the original view, the angle loss is zero although the predicted angles are inaccurate.

To mitigate such a predicted angle ambiguity, we propose a symmetric prior angle (SPA) loss. Based on the SPA loss, the model can be trained to predict precise angles by \textit{indirectly} utilizing the object's symmetric characteristics. As shown in Fig.~\ref{fig:figure4}, the detected objects are symmetric against the symmetry axes (blue-dotted lines) passing the center points of their RBoxes. That is, the pixel contents in the two parts divided by the symmetry axis of the RBox are compared in similarity whose difference is used as supervision for our SPA loss. It is noted that our SPA loss utilizes only symmetric objects, incorporating the symmetry prior from GT class labels for proposals identified as symmetric, such as `airplane,' `ship,' `vehicle,' and `tennis court'

To avoid applying the SPA loss when \(\textit{RB}^{pred}\) are inaccurate for the respective objects, we first check the fidelity scores of proposal, and \textit{sample} the Top-\textit{k} proposals as supervision in the SPA loss as: 
\vspace{-0.2cm}
\begin{equation}\label{eq:12}
\{\textit{RB}^{pred}_n\}^{N_{\text{spa}}}_{n=1} = \text{Top-}k\left(\{\textit{RB}^{pred}_l\}^{N_p}_{l=1} \,|\, \text{sc}^i_{\text{cls}} + \text{sc}^i_{\text{loc}}\right)
\end{equation}
where \(\text{sc}^{(i)}_\text{cls}\) and \(\text{sc}^{(i)}_\text{loc}\) are the classification and localization scores for the \(\textit{l}\)-th \(\textit{RB}^{pred}\), and \(\textit{N}_\textit{p}\) is the total number of proposals. From Eq.~\ref{eq:12}, the selected \(\textit{N}_{spa}\) proposals are considered in the SPA loss by which the predicted angles of \(\textit{RB}^{pred}\) (\(\theta^{pred}_{rb}\)) are enforced to align with the orientations of objects in the sense of maximizing the similarity, Structural Similarity Index (SSIM \cite{ssim}), between the pixel contents in the two parts of each \(\textit{RB}^{pred}\). It should be noted that, even in cases where symmetric objects may not appear perfectly symmetric due to contextual factors like shadows or asymmetrical cargo arrangements, their symmetry is still maintained by the inherent structural symmetry between the two parts. Our SPA loss is defined as:             
\begin{equation}\label{eq:13}
\vspace{-0.1cm}
    \mathcal{L}_{\text{SPA}} = ({1}/{N_{\text{spa}}}) \textstyle\sum_{n=1}^{N_{\text{spa}}} \left(1 - \text{SSIM}(I_{p1}^{(n)}, I_{p2}^{(n)})\right)
\vspace{-0.1cm}
\end{equation}
To remove the influence of object sizes in \(\textit{L}_\text{SPA}\) computation, the proposals (\(\textit{RB}^{pred}\)) are projected onto a fixed-size grid of \(50\times50\). Then, the pixel content (\(\textit{I}_{p1}\)) in one part of the proposal's projection is compared with that (\(\textit{I}_{p2}\)) of the other part that is flipped before the comparison.     

\subsection{Loss Functions}  
\label{subsec:loss functions}

In the orientation learning branch (OLB), two angle-based losses \cite{h2rbox-v2}, \( \mathcal{L}_{\text{rot}} \) and \( \mathcal{L}_{\text{flp}} \), are adopted to leverage the consistency between the original, rotated, and flipped views of each object proposal. For the rotated and flipped views, \( \mathcal{L}_{\text{rot}} \) and \( \mathcal{L}_{\text{flp}} \) are computed by comparing with the predicted angle \( \theta \) in the original view ($\text{I}_\textit{ori}$):
\begin{equation}
        \mathcal{L}_{\text{rot}} = l_{\textit{s}}(\theta_{\text{rot}} - \theta, R),
        { \mathcal{L}}_{\text{flp}} = l_{\textit{s}}(\theta_{\text{flp}} + \theta, 0),
\end{equation}
where \(\textit{l}_{\textit{s}}\) denotes a smooth L1 loss-based snap loss \cite{h2rbox-v2}, and \( R \) denotes the angle applied to $\text{I}_\textit{ori}$. The final angle loss  is:
\begin{equation}
    \mathcal{L}_{\text{ang}} = \beta ({\lambda_r}\mathcal{L}_{\text{rot}} + {\lambda_f}\mathcal{L}_{\text{flp}}) + \gamma \mathcal{L}_{\text{SPA}},
\end{equation}
where \( {\lambda_{r}}=1.0,  {\lambda_{f}}=0.05\), \(\beta=0.6\), and $\gamma=0.05$ are empirically determined for all our experiments.
In the shape learning branch (SLB), we use our IoU-based \cite{iou} regression loss \( \mathcal{L}_{\text{reg}} \) in Eq.~\ref{eq:regression_loss}. The overall loss is defined as:
\begin{equation}
    \mathcal{L}_{\text{total}} = \lambda_{\text{ang}} \mathcal{L}_{\text{ang}} + \lambda_{\text{reg}} \mathcal{L}_{\text{reg}} + \lambda_{\text{cn}} \mathcal{L}_{\text{cn}} + \lambda_{\text{cls}} \mathcal{L}_{\text{cls}}
\end{equation}
where \( \mathcal{L}_{\text{cn}} \) is the center-ness loss \cite{fcos} , and \( \mathcal{L}_{\text{cls}} \) is the classification loss based on the focal loss \cite{retinanet}. The weighting factors, \( \lambda_{\text{ang}} \), \( \lambda_{\text{reg}} \), \( \lambda_{\text{cn}} \), and \( \lambda_{\text{cls}} \) are all set to 1.

\section{Experiments}

\label{sec:Experiments}
\begin{table}[t]
    \scriptsize
    \centering
    \setlength{\tabcolsep}{3pt} 
\begin{tabular}{l|c|c|c|c|c}
\toprule
\multirow{2}{*}{Datasets} & \# of & Image & \# of & \# of & Annotation \\
& Images &  Widths & Objects & Classes & Types \\
\midrule
DIOR \cite{dior} & 22,463 & 800 & 190,288 & 20 &T-HBox \\
DIOR-R \cite{dior-r} & 22,463 & 800 & 190,288 & 20 &RBox \\
DOTA-v1.0 \cite{dota} & 2,806 & 800 $\sim$ 4K & 188,282 &15 & C-HBox, RBox \\
SIMD \cite{simd} & 5,000 & 1024 &45,096&15 & T-HBox \\
NWPU VHR-10 \cite{nwpuvhr-10} & 800 & $\sim$1000 & 3,775 & 10 & T-HBox \\
\bottomrule
\end{tabular}%

    \vspace{-0.2cm}
    \caption{Characteristics of datasets used for experiments}
    \vspace{-0.2cm}
    \label{tab:table1}
\end{table}

\begin{table*}[h]
\scriptsize
\vspace{-0.3cm}
\resizebox{\textwidth}{!}{%
\centering

\setlength{\tabcolsep}{3pt} 
\begin{tabular}{l|l|cccccccccccccccccccc|cc}
\toprule
\multicolumn{2}{c|}{\textbf{Methods}} & \textbf{\underline{APL}} & \textbf{APO} & \textbf{BF} & \textbf{BC} & \textbf{BR} & \textbf{CH} & \textbf{\underline{ESA}} & \textbf{ETS} & \textbf{DAM} & \textbf{GF} & \textbf{GTF} & \textbf{HA} & \textbf{\underline{OP}} & \textbf{SH} & \textbf{STA} & \textbf{STO} & \textbf{TC} & \textbf{TS} & \textbf{VE} & \textbf{WM} & \textbf{\underline{3-AP$_{50}$}} & \textbf{AP$_{50}$} \\ 
\midrule
\multirow{6}{*}{\rotatebox{90}{$\mathcal{S}_R$}} & RetinaNet \cite{retinanet} & 59.8 & 19.3 & 69.7 & 81.3 & 17.2 & 72.7 & 68.7 & 49.4 & 18.4 & 69.5 & 71.3 & 33.3 & 34.1 & 75.8 & 67.1 & 59.6 & 81.0 & 44.1 & 38.0 & 62.5 & 54.20 & 54.64 \\
& FCOS \cite{fcos} & 62.1 & 37.9 & 74.6 & 81.2 & 32.9 & 72.1 & 75.3 & 61.8 & 27.4 & 69.1 & 78.7 & 34.4 & 50.6 & 80.1 & 68.6 & 68.1 & 81.3 & 49.1 & 43.4 & 64.5 & 62.67 & 60.66 \\
& Oriented R-CNN \cite{orientedrcnn} & 63.0 & 36.7 & 71.9 & 81.6 & 41.1 & 72.6 & 77.8 & 65.5 & 24.8 & 72.9 & 82.1 & 40.9 & 56.5 & 81.2 & 73.4 & 62.4 & 81.5 & 53.3 & 43.3 & 65.6 & 65.77 & 62.41 \\
& GWD \cite{gwd} (RetinaNet) & 61.5 & 23.6 & 73.6 & 81.1 & 17.4 & 72.7 & 68.3 & 47.2 & 20.7 & 71.2& 73.2 & 33.9 & 34.3 & 77.6 & 64.7 & 57.5 & 80.9 & 42.1 & 39.7 & 60.2 & 54.70 & 55.07\\
& KLD \cite{kld} (RetinaNet) & 57.8 & 22.6 & 71.5 & 81.2 & 16.9 & 72.7 & 68.9 & 52.1 & 20.6 & 73.5 & 71.0 & 33.7 & 33.2 & 77.1 & 68.9 & 59.9 & 80.9 & 43.9 & 39.1 & 60.9 & 53.30 & 55.32 \\
& KFIoU \cite{kfiou} (RetinaNet) & 60.6 & 36.6 & 73.6 & 80.9 & 27.0 & 72.6 & 73.4 & 56.5 & 25.4 & 73.9 & 72.0 & 32.9 & 45.8 & 75.8 & 65.2 & 57.6 & 80.0 & 48.0 & 40.1 & 58.8 & 59.93 & 57.84 \\
\midrule
\multirow{1}{*}{\rotatebox{90}{$\mathcal{S}_I$}} & WSODet\textsuperscript{†} \cite{wsodet} & 20.7 & 29.0 & 63.2 & 67.3 & 0.2 & 65.5 & 0.4 & 0.1 & 0.3 & 49.0 & 28.9 & 0.3 & 1.5 & 1.2 & 53.4 & 16.4 & 40.0 & 0.1 & 6.1 & 0.1 & 7.53 & 22.20 \\
\midrule
\multirow{2}{*}{\rotatebox{90}{$\mathcal{S}_P$}} & PointOBB\textsuperscript{†} \cite{pointobb} & 58.2 & 15.3 & 70.5 & 78.6 & 0.1 & 72.2 & 69.6 & 1.8 & 3.7 & 0.3 & 77.3 & 16.7 & 40.4 & 79.2 & 39.6 & 32.4 & 29.6 & 16.8 & 33.6 & 27.7 & 56.07 & 38.08 \\
& Point2RBox-SK \cite{point2rbox} & 41.9 & 9.1 & 62.9 & 52.8 & 10.8 & 72.2 & 3.0 & 43.9 & 5.5 & 9.7 & 25.1 & 9.1 & 21.0 & 24.0 & 20.4 & 25.1 & 71.7 & 4.5 & 16.1 & 16.3 & 21.97 & 27.26 \\
\midrule
\multirow{3}{*}{\rotatebox{90}{$\mathcal{S}_H$}}& H2RBox \cite{h2rbox} & 57.1 & 14.4 & 72.2 & 82.6 & 17.5 & 71.2 & 56.5 & 55.2 & 14 & 67.7 & 77.9 & 31 & 40.7 & 76.3 & 66.2 & 63.4 & 81.5 & 50.4 & 38 & 57.6 & 51.43 & 54.57 \\
& H2RBox-v2 \cite{h2rbox-v2} & 55.5 & 17.8 & 76.9 & 80.5 & 27.7 & 72.2 & 63.0 & 58.6 & 24.4 & 73.9 & 80.3 & 33.9 & 47.2 & 77.4 & 58.7 & 60.9 & 81.4 & 48.1 & 41.1 & 53.9 & 55.23 & 56.67 \\
\rowcolor{gray!30} & ABBSPO (Ours) & 69.5 & 15.7 & 76.2 & 87.5 & 29.9 & 72.3 & 75.3 & 61.2 & 28.1 & 74.1 & 81.7 & 34.7 & 48.2 & 79.3 & 67.4 & 61.4 & 81.5 & 54.7 & 41.5 & 53.8 & \textbf{64.33} & \textbf{59.70} \\
\bottomrule
\end{tabular}%

}
\vspace{-0.3cm}
\caption{Quantitative results of each category on the DIOR-R \cite{dior-r} test dataset for RBox-supervised ($\mathcal{S}_R$), Image-supervised ($\mathcal{S}_I$), Point-supervised ($\mathcal{S}_P$) and HBox-supervised ($\mathcal{S}_H$) methods. 
The \underline{3-AP$_{50}$} represents the mean AP$_{50}$ scores for three complex-shaped object categories: `airplane' (APL), `expressway service area' (ESA), and `overpass' (OP). The notation † indicates its results in the paper \cite{pointobb}.}
\label{tab:table2}
\end{table*}

\begin{table*}
\scriptsize
\vspace{-0.1cm}
\resizebox{\textwidth}{!}{%
\centering

\begin{tabular}{l|l|ccccccccccccccc|cc}
\toprule
\multicolumn{2}{c|}{Methods} & \textbf{\underline{PL}} & \textbf{BD} & \textbf{BR} & \textbf{GTF} & \textbf{SV} & \textbf{LV} & \textbf{SH} & \textbf{TC} & \textbf{BC} & \textbf{ST} & \textbf{SBF} & \textbf{RA} & \textbf{HA} & \textbf{\underline{SP}} & \textbf{\underline{HC}} & \textbf{\underline{3-AP$_{50}$}} & \textbf{AP$_{50}$} \\
\midrule
\multirow{7}{*}{\rotatebox{90}{$\mathcal{S}_R$}} & RetinaNet \cite{retinanet} & 87.5 & 75.1 & 39.9 & 59.6 & 66.3& 66.3 & 78.2 & 90.5 & 55.0 & 62.7& 47.1 & 63.6 & 59.4 &55.1  &43.0  &  61.87 & 63.3  \\
&FCOS \cite{fcos} & 88.8 & 74.0 & 46.8 & 59.1 & 70.1& 81.4 & 87.7 & 90.7 & 67.7 & 68.3&  60.2& 66.1 & 64.9 & 58.7 & 44.0 &  63.83& 68.6  \\
&Oriented R-CNN \cite{orientedrcnn} & 89.3& 76.1 & 53.8 & 78.7 &68.6 & 84.9 & 89.3 & 90.8 & 74.3 & 62.8& 66.3 & 66.5 &74.7  & 58.6 & 46.8 & 64.90 &  72.1 \\
& Oriented RepPoints\cite{orientedreppoints} & 89.7 & 80.1 & 50.5& 74.4 & 75.0 & 82.0 & 88.7 &  90.4& 64.0 & 70.0 & 45.7 & 60.6 & 73.6 & 60.4 & 42.8 & 64.30 & 69.86  \\
& GWD \cite{gwd} (RetinaNet)& 88.2 & 74.9 & 41.3 & 60.5 & 66.7 & 68.1 & 85.8 & 90.5 & 50.4 & 66.8 & 45.8 & 65.1 & 60.7 & 52.9 & 38.9 & 60.0 & 63.77  \\
& KLD \cite{kld} (RetinaNet)& 88.4 & 75.8 & 41.4 & 60.0 & 66.1 & 68.8 & 84.7 & 90.6 & 56.8 & 60.4 & 50.4 & 70.1 & 60.0 & 50.5 & 45.7 & 61.53 & 64.65 \\
& KFIoU \cite{kfiou} (RetinaNet)& 84.4 & 74.3 & 40.7 & 55.2 & 57.9 & 56.9 & 76.4 & 71.2 & 46.1 & 64.8 & 54.3 & 65.0 & 58.3 & 48.7 & 42.9 & 58.67 & 59.81 \\
\midrule
\multirow{2}{*}{\rotatebox{90}{$\mathcal{S}_P$}} & PointOBB \cite{pointobb}+FCOS  & 32.4& 67.3 & 0.8 & 53.6 &2.3 &9.7  & 18.8 & 0.3 & 9.9 & 12.8& 0.5 & 54.0 & 11.0 & 34.1 & 11.4 &  25.97 &  21.26 \\
&Point2RBox-SK \cite{point2rbox} &50.1 & 63.7 & 1.6 & 44.7 & 23.9& 34.7 &32.7  & 78.8 & 41.2 &32.2 & 2.1 & 34.3 & 20.8 & 42.5 & 7.2 &  33.27 &  34.03 \\
\midrule
\multirow{3}{*}{\rotatebox{90}{$\mathcal{S}_H$}}& H2RBox \cite{h2rbox} & 89.5 & 73.1 & 37.3 & 55.1 & 70.7 & 76.4 & 85.4 & 90.3 & 66.5 & 67.3 & 59.6 & 64.9 & 60.6 & 57.9 & 36.5 & 61.30 & 66.07 \\
&H2RBox-v2 \cite{h2rbox-v2} & 89.4 & 74.8 & 45.4 & 56.0 & 70.3 & 76.6 & 87.9 & 90.5 & 69.3 & 67.5 & 56.7 & 64.7 & 65.3 & 55.5 & 45.5 & 63.47 & 67.69 \\
\rowcolor{gray!30}
&ABBSPO (Ours)  & 89.2 & 75.6 & 47.4 & 52.8 & 70.3 & 77.6 & 88.2 & 90.5 & 67.9 & 66.8 & 68.2 & 66.2 & 71.6 & 55.6 & 51.0 & \textbf{65.27} & \textbf{69.26} \\
\bottomrule
\end{tabular}%

}
\vspace{-0.3cm}
\caption{Quantitative results of each category on the DOTA-v1.0 \cite{dota} validation dataset for $\mathcal{S}_R$, $\mathcal{S}_I$, $\mathcal{S}_P$ and $\mathcal{S}_H$ methods. The \underline{3-AP$_{50}$} represents the mean AP$_{50}$ scores for three complex-shaped object categories: plane (PL), swimming pool (SP), and helicopter (HC). All the methods are re-trained using only train dataset for fair comparison.}
\vspace{-5mm}
\label{tab:table3}
\end{table*}

\subsection{Datasets}
\label{subsec:datasets_and_implementation_details}
We trained and tested all the methods across four different datasets: DIOR \cite{dior, dior-r}, DOTA-v1.0 \cite{dota}, SIMD \cite{simd} and NWPU VHR-10 \cite{nwpuvhr-10}, which are summarized in Table~\ref{tab:table1}. The details for the datasets and results for SIMD and NWPU are described in \textit{Suppl}.

\subsection{Implementation Details}
Our proposed ABBSPO pipeline adopts the FCOS \cite{fcos} detector as the baseline architecture, utilizing a ResNet-50 \cite{resnet} backbone and an FPN \cite{fpn} neck, based on the H2RBox-v2 \cite{h2rbox-v2} framework. 
To ensure fairness, all models are configured with the ResNet-50 \cite{resnet} backbone and trained for 12 epochs on NVIDIA RTX3090 GPUs.

\begin{table*}[h] 
    \centering
    \makebox[\textwidth][l]{ 
    \begin{minipage}[b][3cm][t]{0.32\textwidth}
        \setlength{\tabcolsep}{4pt}
        \scriptsize
        \centering
        \begin{tabular}{cc|cc|c}
\toprule
\multicolumn{2}{c|}{Module} & \multicolumn{2}{c|}{DIOR-R} & {DOTA-v1.0}\\
\midrule
ABBS & SPA & 3-\(AP_{50}\)& \(AP_{50}\)& \(AP_{50}\) \\
\midrule
 & & 55.23 & 56.67 & 67.69 \\
  \checkmark &  & 62.13 & 58.35 & 68.59 \\
   & \checkmark & 58.77 & 58.99 & 69.16 \\
   \rowcolor{gray!30}
  \checkmark & \checkmark & \textbf{64.33}& \textbf{59.70}& \textbf{69.26} \\
\bottomrule
\end{tabular}

        \vspace{-0.2cm}
        \caption{Ablation results on ABBS module and SPA loss (\(\mathcal{L}_\text{SPA}\)).}
        \label{tab:table4}
    \end{minipage}%
    \hspace{0.1cm} 
    \begin{minipage}[b][3cm][t]{0.3\textwidth}
        \scriptsize
        \centering
        \begin{tabular}{cc|cc}
\toprule
\multicolumn{2}{c|}{Sampling} & \multicolumn{2}{c}{DIOR-R}\\
\midrule
\(\mathcal{L}_\text{SPA}\) & Others & 3-\(AP_{50}\)& \(AP_{50}\) \\
\midrule
 & & 61.67 & 58.93 \\
    \rowcolor{gray!30}
  \checkmark &  & \textbf{64.33}& \textbf{59.70}  \\
  & \checkmark & 43.63 & 50.51 \\
  \checkmark & \checkmark & 45.1 & 50.91 \\
\bottomrule
\end{tabular}

        \vspace{-0.2cm}
        \caption{Ablation results on proposal sampling in \(\mathcal{L}_\text{SPA}\) and other components.}
        \label{tab:table5}
    \end{minipage}%
    \hspace{0.1cm}
    \begin{minipage}[b][3cm][t]{0.32\textwidth}
        \setlength{\tabcolsep}{4pt}
        \scriptsize
        \centering
        \begin{tabular}{@{\hspace{0.1cm}}c@{\hspace{0.1cm}} c@{\hspace{0.1cm}} c@{\hspace{0.1cm}}}
        \begin{tabular}{ccc|cc|c}
\toprule
\multicolumn{3}{c|}{Scale Range} & \multicolumn{2}{c|}{DIOR-R} & {DOTA-v1.0} \\
\midrule
Min & Max & Interval &  3-\(AP_{50}\) & \(AP_{50}\) & \(AP_{50}\) \\
\midrule
0.9 & 1.1 & 0.05 & 57.97 & 58.15 & \textbf{69.26} \\
0.5 & 1.5 & 0.1 & 61.67 & 59.62 & 68.8 \\
1.0 & 1.5 & 0.1 & \textbf{64.33} & \textbf{59.70} & 68.9 \\
\rowcolor{white}
1.0 & 2.0 & 0.1 & 56.07 & 55.46 & 66.55 \\
\bottomrule
\end{tabular}

        \end{tabular}
        \vspace{-0.2cm}
        \caption{Ablation results on scale range in ABBS module.}
        \label{tab:table6}
    \end{minipage}
    }
\end{table*}

\begin{figure*} [htbp]
    \centering
    \includegraphics[width=\linewidth]{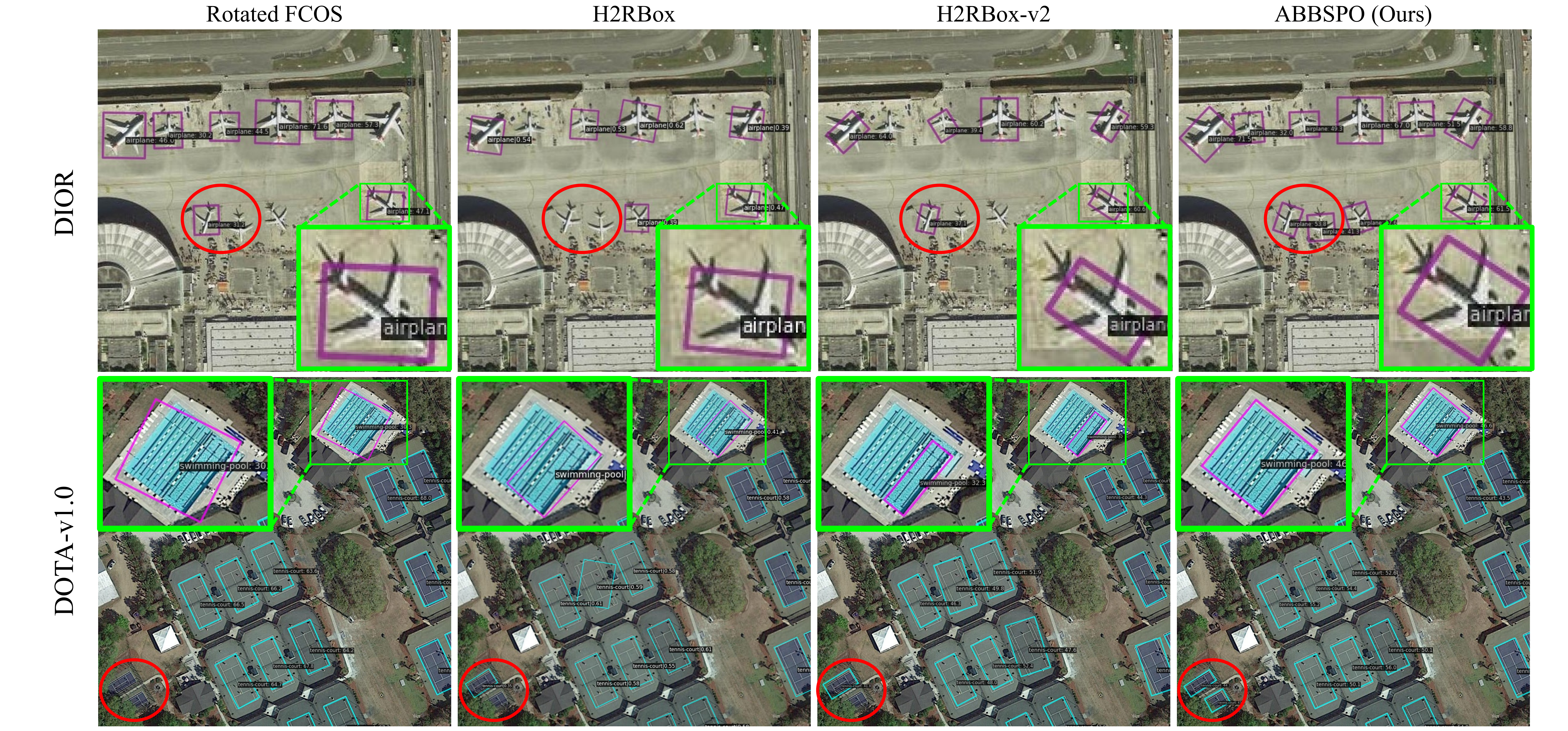}
    \vspace{-0.6cm}
    \caption{Qualitative results on DIOR \cite{dior, dior-r}  and DOTA-v1.0 \cite{dota} . Zoom-in for better visualization. Rotated FCOS was trained only with GT RBoxes, while H2RBox, H2RBox-v2 and our ABBSPO were trained with GT T-HBoxes (1st row) and GT C-HBoxes (2nd row).}
    \vspace{-0.4cm}
    \label{fig:figure5}
\end{figure*}

\subsection{Experimental Results}
\subsubsection{Quantitative Comparison}
\label{subsubsec:quantitative result}
It should be noted that objects such as round-shaped pools have \textit{orientation ambiguities} regardless of their annotations (RBoxes) \cite{h2rbox}. In order to avoid confusion in orientation learning, annotations are modified as having horizontal orientations if the objects belong to the following categories: (i) \textbf{DIOR-R}: `baseball field', `chimney', `golf field', `stadium', `storage tank', `windmill'; and (ii) \textbf{DOTA-v1.0}: `baseball diamond', `stadium', `roundabout'. Accordingly, their orientation learning is enforced to predict the horizontal orientations, similar to previous works \cite{h2rbox, h2rbox-v2}. 

\vspace{-0.2cm}
\noindent\textbf{Results on DIOR-R.} Table~\ref{tab:table2} shows the OOD results. 
In addition to \(\text{AP}_\text{50}\) metric, we use 3-\(\text{AP}_\text{50}\) that focuses on the detection performance of the three complex-shaped object categories: `airplane' (APL), `expressway service area' (ESA), and `overpass' (OP). As shown, our ABBSPO outperforms all weakly supervised OOD methods. Especially, in terms of 3-\(\text{AP}_\text{50}\), our ABBSPO is superior to the HBox-supervised SOTA methods, H2RBox and H2RBox-v2, with large margins of average \textbf{12.9\%}-point and average \textbf{9.1\%}-point improvements. In overall \(\text{AP}_\text{50}\) performance, our ABBSPO surpasses H2RBox by \textbf{5.13\%}-point and the H2RBox-v2 by \textbf{3.03\%}-point. It is noted that our ABBSPO not only surpasses our base detector (H2RBox-v2 \cite{h2rbox-v2}) but also performs comparably to other RBox-supervised OOD methods, such as FCOS \cite{fcos} and Oriented R-CNN \cite{orientedrcnn}. It is worth noting that, compared to the RBox-supervised OOD methods, our ABBSPO shows \textit{even} superior performance with \textit{large} margins from \textbf{6.5}\%-point to \textbf{11.7}\%-point, especially on the `airplane' that has the most complex shape.
Notably, the ABBS module is less effective for rectangular objects, such as ‘tennis court’ (TC) and ‘vehicle’ (VE), as scaling is often unnecessary.
However, it proves highly beneficial for complex-shaped objects, such as the ESA.
The SPA loss is applied only to symmetric categories and helps improve their performance, except for the categories with \textit{orientation ambiguities}, such as `storage tank' (STO).
Since the predicted angles learned through the SPA loss are also utilized in the ABBS module for scale adjustment, both the SPA loss and the ABBS module jointly contribute to performance improvement in symmetric categories. 
This joint effect is particularly evident in complex-shaped symmetric categories, such as APL and ESA, where performance gains are more significant.
Nevertheless, the performance gains for the two symmetric and rectangular categories, TC and VE, are marginal. This is mainly because the ABBS module has limited impact on rectangular shapes, and the small object sizes lead to an insufficient number of pixels for reliably determining the symmetry axis via the SPA loss.\\
\noindent\textbf{Results on DOTA-v1.0.} Table~\ref{tab:table3} shows the detection performance results on the DOTA-v1.0 \cite{dota}. Due to the non-responsiveness of the DOTA evaluation server, we report our experimental results on the validation dataset (458 images) instead of the test dataset (937 images). It should be noted that the validation dataset was not used for training all the methods for fair comparison. We use 3-\(\text{AP}_\text{50}\) that measures the detection performance for the three complex-shaped object categories: `plane', `swimming pool' and `helicopter'.
Our ABBSPO achieves SOTA performance, outperforming H2RBox by \textbf{3.19\%}-point and H2RBox-v2 by \textbf
{1.57\%}-point improvements. Moreover, our ABBSPO even surpasses the FCOS Baseline by \textbf{0.66\%}-point lift.
\vspace{-0.3cm}
\subsubsection{Qualitative Comparison}
\noindent\textbf{Results on DIOR.} As shown in the figures of the first row in Fig.~\ref{fig:figure5}, our ABBSPO is the only method that accurately captures both the orientation and scale of the airplane. Since DIOR annotations provide GT in T-HBox format, direct usage of T-HBoxes as GT for training to predict RBox leads to degradations in orientation and scale prediction accuracy for the existing HBox-supervised OOD methods as shown in the figures of columns 2 and 3 in Fig.~\ref{fig:figure5}. In contrast, our ABBSPO avoids such degradation by utilizing the ABBS module that optimally scales the GT HBox sizes for precise RBox prediction during training. It is also worthwhile to mention that the predicted orientations by our ABBSPO are more precisely obtained via our SPA loss. Furthermore, it should be noted that compared to the RBox-supervised baseline method (Rotated FCOS \cite{fcos}), our approach demonstrates superior visual results, even under weakly supervised learning.\\
\noindent\textbf{Results on DOTA-v1.0.} As shown in the figures of the second row in Fig.~\ref{fig:figure5}, ABBSPO very accurately predicts both the orientation and scale of the swimming pool, achieving similar accuracy for tennis court. Interestingly, only ABBSPO successfully detects the two tennis courts that are partially occluded by trees (red solid circle) while the other methods failed. These results visually support the effectiveness of our ABBS module and SPA loss in learning the scales and orientations of objects accurately.
\subsection{Ablation Studies}
\noindent\textbf{Ablation study on SPA loss and ABBS module.} As shown in Table~\ref{tab:table4}, both components contribute to performance improvements. The ABBS module effectively scales the GT HBoxes, leading to an increase in \(\text{AP}_{50}\) performance on the DIOR dataset. 
Notably, it has a greater effect on complex-shaped object categories, resulting in a significant improvement in 3-\(\text{AP}_\text{50}\).
Similarly, the SPA loss enhances angle prediction accuracy, also bringing an improvement in \(\text{AP}_{50}\). \\
\noindent\textbf{Ablation study on proposal sampling.} As shown in Table~\ref{tab:table5}, applying Top-\textit{k} proposal sampling exclusively to the SPA loss (\(\mathcal{L}_\text{SPA}\)) yields the highest \(\text{AP}_{50}\) performance, as the symmetric proposals of high-quality benefits \(\mathcal{L}_\text{SPA}\). But, additional proposal sampling to the others (\(\mathcal{L}_\text{rot}\), \(\mathcal{L}_\text{flp}\), \(\mathcal{L}_\text{reg}\), \(\mathcal{L}_\text{cn}\), \(\mathcal{L}_\text{cls}\)) significantly lowers the performance.\\
\noindent\textbf{Ablation study on scale range in the ABBS module.} As shown in Table~\ref{tab:table6}, the optimal scale range is influenced by the type of GT HBoxes. For DIOR’s T-HBoxes, a scale range of 1 to 1.5 works well because it ensures that the predicted RBoxes fully cover the objects boundary. On the other hand, for DOTA’s C-HBoxes, which are already close to the optimal HBoxes, the optimal scale range is closer to 1. By adjusting the scale range based on the type of HBoxes, the ABBS module achieves high accuracy in predicting RBoxes for both datasets.

\section{Conclusion}
\label{sec:conclusion}
Our ABBSPO, a weakly supervised OOD framework, effectively learns RBox prediction regardless of the type of HBox annotations (T-HBox and C-HBox). With our proposed Adaptive Bounding Box Scaling (ABBS) and Symmetric Prior Angle (SPA) loss, we achieved enhanced orientation and scale accuracy for OOD, which is comparable to or even better than RBox-supervised methods.
Extensive experimental results underscore the superiority of our approach, surpassing state-of-the-art HBox-supervised methods. Our method effectively bridges the gap between weakly supervised OOD and fully supervised OOD, making it a promising solution for applications requiring efficient and accurate object detection via training with relatively cheap annotations of HBoxes compared to RBoxes.

{
    \small
    \bibliographystyle{ieeenat_fullname}
    \bibliography{main}
}
\clearpage

\clearpage
\setcounter{page}{1}
\twocolumn[{
\renewcommand\twocolumn[1][]{#1}%
    \begin{center}
           \includegraphics[width=\linewidth]{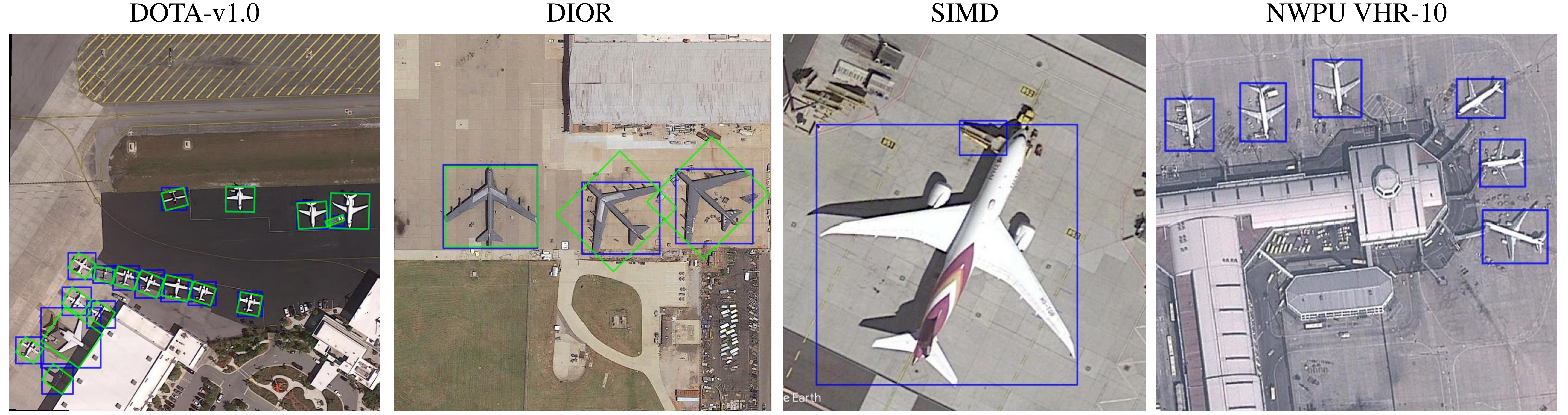}
           \captionof{figure}{Visualization of GT Box Types for Different Datasets.}
    \label{fig:figure6}
    \end{center}
}]

\appendix

\begin{figure*} [htbp]
    \centering
    \includegraphics[width=\linewidth]{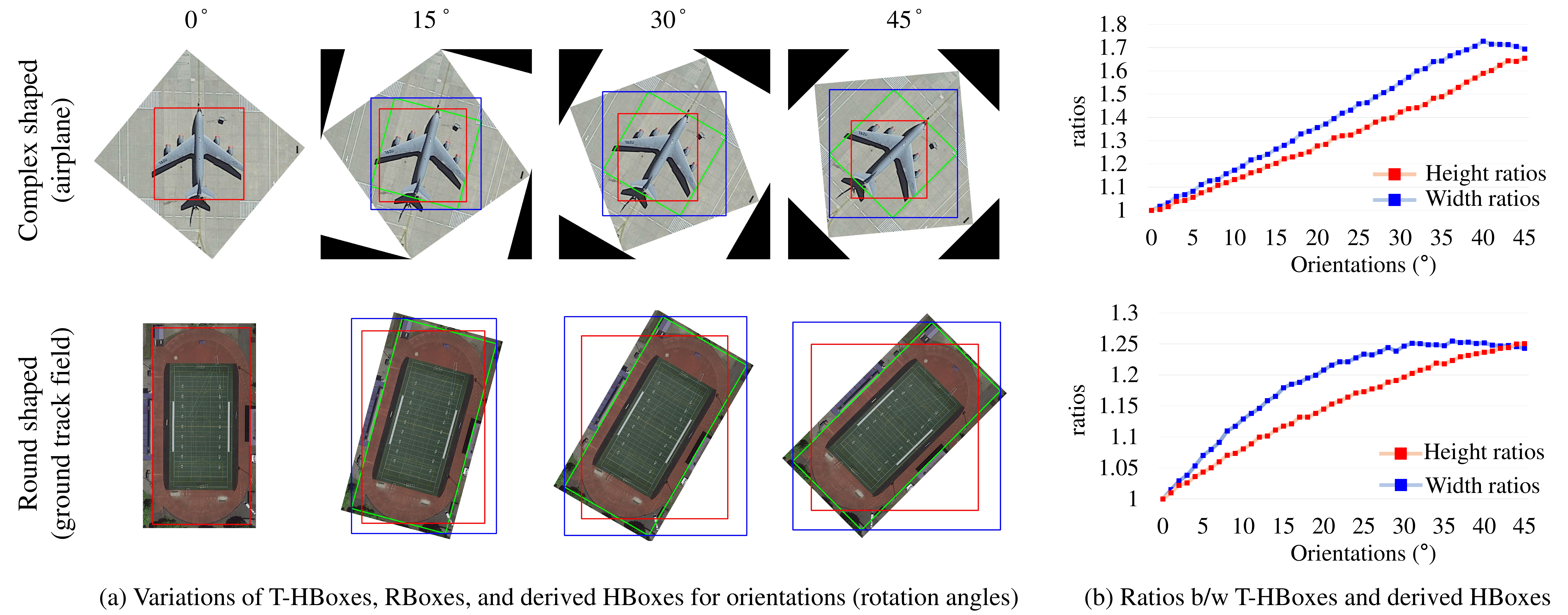}
    \caption{ Analysis on variations of width and height ratios between the T-HBoxes and the HBoxes derived as minimum circumscribed rectangles of RBoxes. (a) Variations in the shapes of manually annotated T-HBoxes, manually annotated RBoxes and derived HBoxes from the RBoxes for various rotations angles of different objects (airplanes in the first row and ground track fields in the second row); (b) Width and height ratios between T-HBoxes and corresponding derived HBoxes for the rotation angles of the airplanes (top) and the ground track fields (bottom).}
    \label{fig:figure7}
\end{figure*}

\begin{figure*} [htbp]
    \centering
    \includegraphics[width=\linewidth]{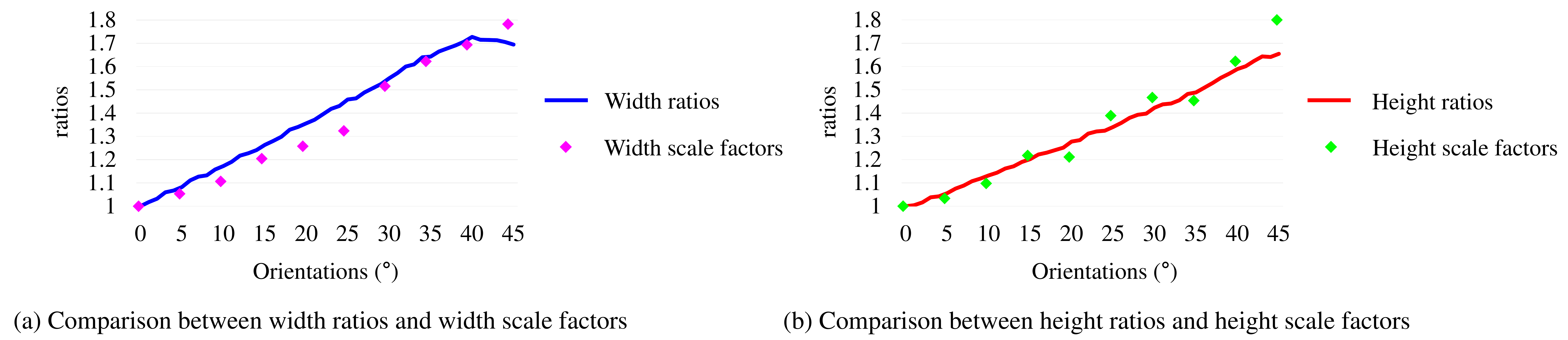}
        \caption{Comparisons between width (height) ratios and scale factors. (a) Width ratio curve (\textcolor{blue}{blue curve}) calculated from manual annotations, and \textit{angle-adjusted} width scale factors (\textcolor{violetred}{diamond-shaped pink points}) derived from best scaled HBoxes selected by the ABBS; (b) height ratio curve (\textcolor{red}{red curve}) calculated from manual annotations and \textit{angle-adjusted} height scale factors (\textcolor{green}{diamond-shaped green points}) derived from best scaled HBoxes selected by the ABBS. It is noted that the \textit{angle-adjusted} width (height) scale factors are well aligned with the width (height) curves, indicating that our ABBS works properly in scaling the GT HBoxes.}    
    \label{fig:figure8}
\end{figure*}
\section{Details of Datasets}
\label{sec:details of datasets}
\subsection{Detailed Description of Datasets}

\noindent\textbf{DOTA-v1.0} \cite{dota} comprises 2,806 images with 188,282 annotated instances across 15 categories, having both rotated bounding boxes (GT RBoxes) and coarse horizontal bounding boxes (GT C-HBoxes) for annotation. Among these images, 1,411 are designated for training, 458 for validation, and 937 for testing. The image dimensions range from \(800\times800\) to \(4,000\times4,000\). During training, the images were cropped into a patch size of \(1,024 \times 1,024\). The dataset includes 15 categories: `plane' (PL), `baseball-diamond' (BD), `bridge' (BR), `ground-track-field' (GTF), `small-vehicle' (SV), `large-vehicle' (LV), `ship' (SH), `tennis-court' (TC), `basketball-court' (BC), `storage-tank' (ST), `soccer-ball-field' (SBF), `roundabout' (RA), `harbor' (HA), `swimming-pool' (SP), and `helicopter' (HC). 

In our experimental setup, all methods including our ABSSPO are trained on the train split of DOTA-v1.0 \cite{dota} with GT C-HBoxes and evaluated on its validation split with GT RBoxes.

\noindent\textbf{DIOR} \cite{dior} contains \(800\times800\)-sized 23,463 aerial images of 20 categories with 190,288 instances (objects), each having tight horizontal bounding box annotations (GT T-HBoxes). Among these, 5,862 images are used for training, 5,863 images for validation, and the remaining 11,738 images for testing. The 20 categories of the dataset include: `airplane' (APL), `airport' (APO), `baseball field' (BF), `basketball court' (BC), `bridge' (BR), `chimney' (CH), `expressway service area' (ESA), `expressway toll station' (ETS), `dam' (DAM), `golf field' (GF), `ground track field' (GTF), `harbor' (HA), `overpass' (OP), `ship' (SH), `stadium' (STA), `storage tank' (STO), `tennis court' (TC), `train station' (TS), `vehicle' (VE), and `windmill' (WM).

\noindent\textbf{DIOR-R} \cite{dior-r} contains the same images as DIOR \cite{dior}, but includes rotated bounding box annotations for its objects instead of HBox annotations. It should be noted that in our experiments, we utilize both the train and validation splits in DIOR \cite{dior} dataset for training our method, while employing the test split in DIOR-R \cite{dior-r} 
 dataset for evaluation.

\noindent\textbf{SIMD} \cite{simd} comprises aerial images annotated with tight horizontal bounding boxes (GT T-HBoxes). The dataset contains 5,000 images spanning 15 categories, with a total of 45,096 instances. Each image has a fixed width of 1,024 pixels and fixed-sized heights of 768 pixels. The 15 categories are as follows: `car', `truck', `van', `longvehicle', `bus', `airliner', `propeller', `trainer', `chartered', `fighter', `other', `stairtruck', `pushbacktruck', `helicopter', and `boat'.

\noindent\textbf{NWPU VHR-10} \cite{nwpuvhr-10} is an aerial image dataset featuring tight horizontal bounding box (GT T-HBox) annotations. It comprises 800 images spanning 10 categories, with approximately 3,775 annotated instances. The images have widths of around 1,000 pixels. The dataset includes the following 10 categories: `airplane', `ship', `storage tank', `baseball diamond', `tennis court', `basketball court', `ground track field', `harbor', `bridge', and `vehicle'.

After training on the SIMD \cite{simd} and NWPU VHR-10 \cite{nwpuvhr-10} datasets, we focus on qualitative comparisons only, as these datasets do not provide RBox annotations.

\subsection{Visualization of GT Box Types for Different Datasets.}
Fig.~\ref{fig:figure6} illustrates how GT HBoxes are annotated across different datasets. In this figure, the GT RBoxes are marked in \textcolor{green}{green}, while the GT HBoxes are marked in \textcolor{blue}{blue}. For the SIMD \cite{simd} and NWPU VHR-10 datasets \cite{nwpuvhr-10} where their GT RBoxes do not exist, only the GT HBoxes (\textcolor{blue}{blue}) are displayed. The comparison emphasizes airplanes that have complex shapes, to highlight the differences in the GT HBox annotation types.

\noindent\textbf{DOTA-v1.0}  \cite{dota}. The GT HBoxes (\textcolor{blue}{blue}) in DOTA are annotated as the minimum circumscribed HBoxes for their corresponding GT RBoxes (\textcolor{green}{green}), as can be seen in the first column of Fig.~\ref{fig:figure6}. For the objects with larger rotation angles, the sizes of their GT HBoxes (\textcolor{blue}{blue}) appear larger-sized, when being more apart from the objects' boundaries. Such GT HBoxes (\textcolor{blue}{blue}) that are derived directly from the objects boundaries were previously defined as GT C-HBoxes in Sec.~\ref{sec:intro}.\\
\noindent\textbf{DIOR} \cite{dior} (\& \textbf{DIOR-R} \cite{dior-r}). Its GT HBoxes (\textcolor{blue}{blue}) are sourced from the DIOR dataset, while the GT RBoxes (\textcolor{green}{green}) are taken from the DIOR-R dataset for visualization purposes. The GT HBoxes (\textcolor{blue}{blue}) tightly enclose the objects' boundaries, and are annotated independently from the GT RBoxes (\textcolor{green}{green}). This type of HBox annotations is referred to as GT T-HBoxes. As the orientations of the objects increase, their GT RBoxes (\textcolor{green}{green}) extend further beyond their corresponding GT HBoxes (\textcolor{blue}{blue}), which can be observed in the center and right airplanes in the second column of Fig.~\ref{fig:figure6}.\\
\noindent\textbf{DOTA} \cite{dota} vs. \textbf{DIOR} \cite{dior} (\& \textbf{DIOR-R} \cite{dior-r}).  In the DOTA and DIOR datasets, their objects are annotated as GT C-HBoxes and GT T-HBoxes, respectively. The GT HBoxes (\textcolor{blue}{blue}) and GT RBoxes (\textcolor{green}{green}) align perfectly each other when the objects' orientations are horizontal or vertical, as demonstrated by the top-right airplanes in the first column of Fig.~\ref{fig:figure6} and the left airplane in the second column of Fig.~\ref{fig:figure6}. However, as the objects get more rotated from the horizontal or vertical angle, their GT C-HBoxes (\textcolor{blue}{blue}) become consistently enlarged to circumscribe their corresponding GT RBoxes (\textcolor{green}{green}), as seen in the bottom-left airplanes in the first column of Fig.~\ref{fig:figure6}. In contrast, as depicted in the second column of Fig.~\ref{fig:figure6}, the GT T-HBoxes always tightly enclose the objects boundaries, regardless of their GT RBoxes (\textcolor{green}{green}). This difference in GT HBox annotation leads to a significant degradation in the OOD performance of the previous HBox-supervised OOD methods when trained on the DIOR dataset with GT T-HBoxes, unlike when trained on the DOTA dataset with GT C-HBoxes. \\
\noindent\textbf{SIMD} \cite{simd} \& \textbf{NWPU VHR-10} \cite{nwpuvhr-10}. As shown in the third and forth columns of Fig.~\ref{fig:figure6}, SIMD and NWPU VHR-10 datasets contains only GT T-Hboxes, where the GT T-HBoxes (\textcolor{blue}{blue}) tightly enclose the boundaries of airplanes, even for objects with large orientation angles.
\section{Additional Ablation Study}
\label{additional ablation study}
\noindent\textbf{Ablation Study on Scale Adjustment Function.}
\noindent As mentioned in Sec.~\ref{subsec:adaptive_bounding_box_scaling_module}, we incorporate the object shape types and orientation degrees into the scale adjustment of the widths and heights of GT T-HBoxes. The scale adjustment function \textit{f} is designed as a linear function of the angle $\theta$, as presented in Eq.~\ref{eq:8}. \\
Fig.~\ref{fig:figure7}-(a) demonstrates this process using two images containing objects from the `airplane' class and `ground track field' class, which are manually rotated from 0° to 45° in 15° increments. On the top of each rotated image, a T-HBox (\textcolor{red}{red}), an RBox (\textcolor{green}{green}), and a minimum circumscribed HBox (\textcolor{blue}{blue}) derived from the RBox are overlayed.\\
Fig.~\ref{fig:figure7}-(b) shows the width and height ratios between the T-HBoxes and their derived HBoxes. As shown, the results indicate that these ratios increase linearly as the orientation degrees increase, validating that the proposed linear scale adjustment function \textit{f} can effectively capture the scale variations in annotations caused by the object orientations. \\
Additionally, Fig.~\ref{fig:figure8} visualizes the scale proportion between the annotated T-HBoxes and their best scaled HBoxes in the ABBS module, by passing manually rotated airplane images from 0° to 45°. This scale proportion corresponds to the \textit{angle-adjusted} scale factors derived from Eq.~\ref{eq:7}, Eq.~\ref{eq:8}, and Eq.~\ref{eq:abbs_loss}, which are represented as diamond-shaped points in Fig.~\ref{fig:figure8}. Fig.~\ref{fig:figure8}-(a) compares the width ratio curve (\textcolor{blue}{blue curve}) calculated from manually annotated T-HBoxes and derived HBoxes from manually annotated RBoxes with the \textit{angle-adjusted} width scale factors (\textcolor{violetred}{diamond-shaped pink points}) obtained from best scaled HBoxes selected by the ABBS. Fig.~\ref{fig:figure8}-(b) compares the height ratio curve (\textcolor{red}{red curve}) calculated from manually annotated T-HBoxes and derived HBoxes from manually annotated RBoxes with the \textit{angle-adjusted} height scale factors (\textcolor{green}{diamond-shaped green points}) obtained from best scaled HBoxes selected by the ABBS. It is noted in Fig.~\ref{fig:figure8}-(a) and -(b) that the \textit{angle-adjusted} width (height) scale factors are well aligned with the width (height) curves, indicating that our ABBS works properly in scaling the GT HBoxes. This alignment highlights the effectiveness of our proposed ABBS module in capturing and leveraging the scale variations caused by the object orientation, enabling precise adaptation to changes in object orientations and shapes during training.

\begin{table}[t] 
\centering
\begin{tabular}{ccc|cc}
\toprule
\multicolumn{3}{c|}{Scale Range} & \multicolumn{2}{c}{DOTA-v1.0}\\
\midrule
Min & Max & Interval &  3-\(AP_{50}\) & \(AP_{50}\)  \\
\midrule
0.9 & 1.0 & 0.05 & 61.80 & 68.09  \\
1.0 & 1.1 & 0.05 & 64.77 & 69.08 \\
\rowcolor{gray!30}
0.9 & 1.1 & 0.05 & \textbf{65.27} & \textbf{69.26} \\
\rowcolor{white}
0.8 & 1.1 & 0.05 & 63.33 & 69.03 \\
\bottomrule
\end{tabular}

\caption{Ablation results on the scale ranges of GT HBoxes for ABBS module in the DOTA-v1.0 \cite{dota} dataset.}
\label{tab:table7}
\end{table}

\begin{table*}[h]
\scriptsize
\resizebox{\textwidth}{!}{%
\centering
\setlength{\tabcolsep}{3pt} 
\begin{tabular}{l|cccccccccccccccccccc|cc}
\toprule
\multicolumn{1}{c|}{\textbf{Methods}} & \textbf{\underline{APL}} & \textbf{APO} & \textbf{BF} & \textbf{BC} & \textbf{BR} & \textbf{CH} & \textbf{\underline{ESA}} & \textbf{ETS} & \textbf{DAM} & \textbf{GF} & \textbf{GTF} & \textbf{HA} & \textbf{\underline{OP}} & \textbf{SH} & \textbf{STA} & \textbf{STO} & \textbf{TC} & \textbf{TS} & \textbf{VE} & \textbf{WM} & \textbf{\underline{3-AP$_{50}$}} & \textbf{AP$_{50}$} \\ 
\midrule
H2RBox \cite{h2rbox} & 57.1 & 14.4 & 72.2 & 82.6 & 17.5 & 71.2 & 56.5 & 55.2 & 14 & 67.7 & 77.9 & 31 & 40.7 & 76.3 & 66.2 & 63.4 & 81.5 & 50.4 & 38 & 57.6 & 51.43 & 54.57 \\
H2RBox\textsuperscript{*} \cite{h2rbox} & 65.5 & 12.5 & 74.6 & 81.3& 21.3 & 72.2 & 62.7 & 60.4 & 19.2 & 70.1 & 78.7 & 35.3 & 44.3 & 79.1 & 62.7 & 68.2 & 81.5 & 51.7 & 39.6 & 60.7 & 57.50 & 57.08 \\

H2RBox-v2 \cite{h2rbox-v2} & 55.5 & 17.8 & 76.9 & 80.5 & 27.7 & 72.2 & 63.0 & 58.6 & 24.4 & 73.9 & 80.3 & 33.9 & 47.2 & 77.4 & 58.7 & 60.9 & 81.4 & 48.1 & 41.1 & 53.9 & 55.23 & 56.67 \\
H2RBox-v2\textsuperscript{*} \cite{h2rbox-v2} & 67.2 & 11.5 & 75.8 & 84.0 & 31.4 & 72.5 & 65.3 & 60.7 & 25.3 & 72.2 & 80.9 & 35.2 & 50.2 & 78.9 & 67.0 & 61.5 & 81.5 & 52.6 & 43.0 & 26.7 & 60.90 & 57.17\\

\rowcolor{gray!30}
ABBSPO (Ours) & 69.5 & 15.7 & 76.2 & 87.5 & 29.9 & 72.3 & 75.3 & 61.2 & 28.1 & 74.1 & 81.7 & 34.7 & 48.2 & 79.3 & 67.4 & 61.4 & 81.5 & 54.7 & 41.5 & 53.8 & \textbf{64.33} & \textbf{59.70} \\
\rowcolor{gray!30}
ABBSPO (Ours)\textsuperscript{*} & 66.6 & 20.2 & 77.6 & 84.7 & 30.8 & 72.5 & 75.0 & 60.1 &28.3 & 75.3 & 81.2 & 35.9  & 49.0 & 79.4 & 69.7 & 65.3 & 81.4 & 55.1 & 41.7& 33.0 & \textbf{63.53} & \textbf{59.14} \\
\bottomrule
\end{tabular}%

}
\caption{Quantitative OOD results for various object categories on the DIOR-R \cite{dior-r} dataset. Results marked with * are obtained by running H2RBox \cite{h2rbox}'s public source codes. For these results, the experiment settings do not include angle prediction for the airplane class. Results without * are obtained using the original configuration.}
\label{tab:table8}
\end{table*}
\begin{figure*} [htbp]
    \centering
    \includegraphics[width=\linewidth]{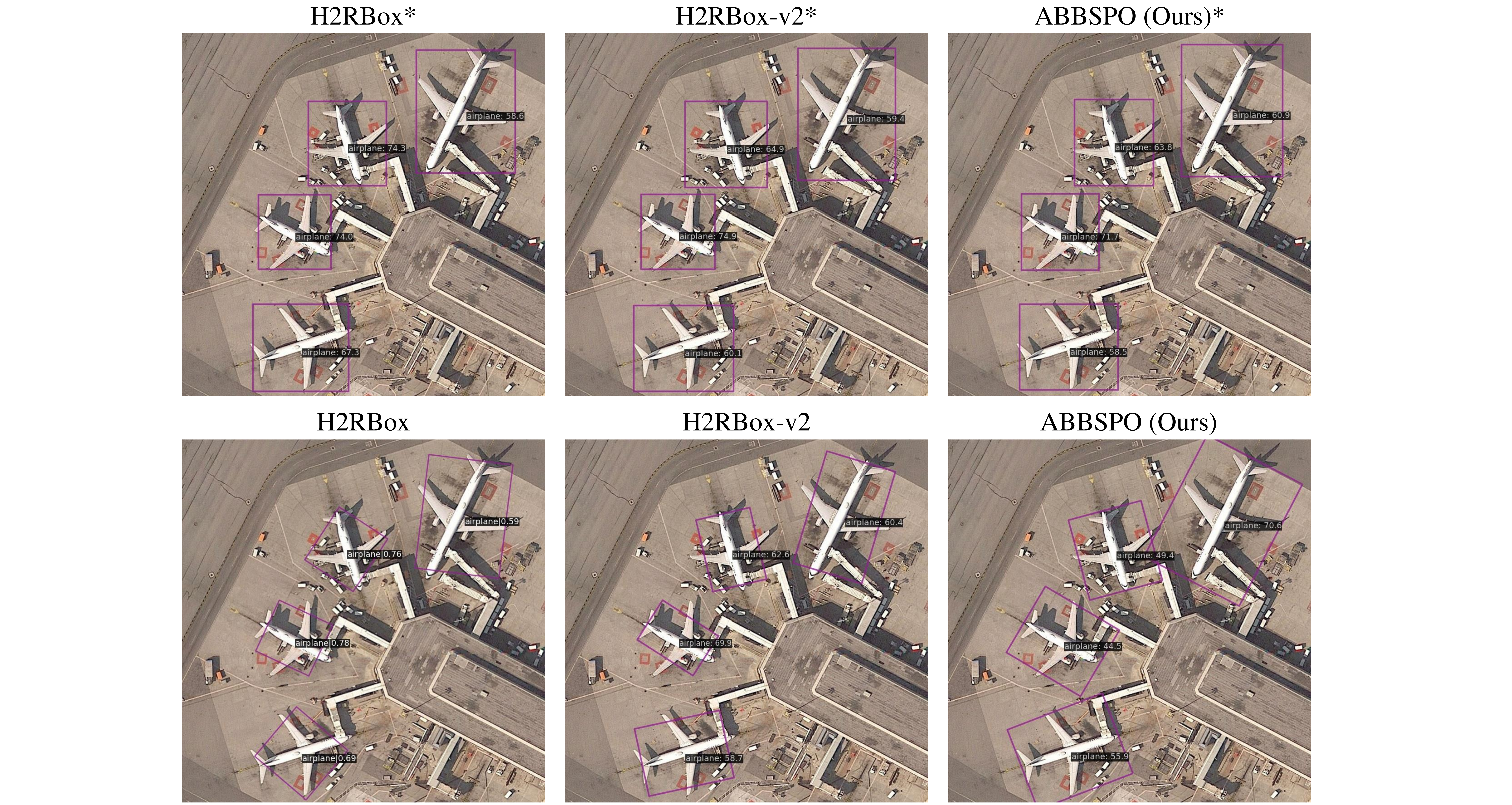}
    \caption{Qualitative OOD comparisons on DIOR \cite{dior, dior-r}. Results marked with * are obtained by running H2RBox \cite{h2rbox}'s public source codes. For these results, the experiment settings do not include angle prediction for the airplane category. Results without * are obtained using the original configuration.}
    \label{fig:figure9}
\end{figure*}
\noindent\textbf{Ablation Study on Scale Ranges of GT HBoxes for ABBS Module.}
\noindent Our ABBS module adjusts the sizes of given GT HBoxes during the training, with their scale adjustment range depending on the annotation types of the datasets.
Especially for the DIOR dataset \cite{dior} that uses GT T-HBoxes, the scale range is set from 1 to 1.5 to optimize the training process as shown in Table~\ref{tab:table6}. Conversely, for the DOTA-v1.0 dataset \cite{dota} that provides GT C-HBoxes, a narrower scale range of 0.9 to 1.1 is employed. 
As shown in Table~\ref{tab:table7}, the best OOD performance on DOTA-v1.0 dataset is achieved when the scale range is set between 0.9 and 1.1, validating the necessity of dataset-specific scale adjustments.

\section{Additional Results on DIOR}
\label{additional quantitative results}
As discussed in Sec.~\ref{subsubsec:quantitative result}, our ABBSPO demonstrates superior OOD performance compared to the previous methods, H2RBox \cite{h2rbox} and H2RBox-v2 \cite{h2rbox-v2}. 
For fair comparison, we set the objects belonging to the following six classes (‘baseball field’, ‘chimney’, ‘golf field’, ‘stadium’, ‘storage tank’, and ‘windmill’) as the subjects not to predict their orientations due to orientation ambiguities. 
We denote this setting as the original configuration. We further provide the OOD results under an additional configuration, where six different classes (‘airplane’, ‘baseball field’, ‘chimney’, ‘golf field’, ‘stadium’, and ‘storage tank’) were designated as objects without orientation prediction, following the H2Rox \cite{h2rbox}'s public source codes.

We denote the methods trained and evaluated in the above additional configuration as H2RBox\textsuperscript{*} \cite{h2rbox}, H2RBox-v2\textsuperscript{*} \cite{h2rbox-v2}, and ABBSPO\textsuperscript{*}, while H2RBox \cite{h2rbox}, H2RBox-v2 \cite{h2rbox-v2}, and ABBSPO denote the methods trained and evaluated in the original configuration.
As illustrated in Fig.~\ref{fig:figure9}, H2RBox\textsuperscript{*} \cite{h2rbox}, H2RBox-v2\textsuperscript{*} \cite{h2rbox-v2}, and ABBSPO\textsuperscript{*} predict objects in ‘airplane’ class in the form of HBoxes without orientations, while H2RBox \cite{h2rbox}, H2RBox-v2 \cite{h2rbox-v2}, and ABBSPO predicts orientations of the objects for the same class.
As shown in Table~\ref{tab:table8}, our ABBSPO\textsuperscript{*} still outperforms H2RBox\textsuperscript{*} \cite{h2rbox} and H2RBox-v2\textsuperscript{*} \cite{h2rbox-v2}.
These findings demonstrate the effectiveness and robustness of our ABBSPO over different configurations for the orientation ambiguity.

\begin{figure*} [htbp]
    \centering
    \includegraphics[width=\linewidth]{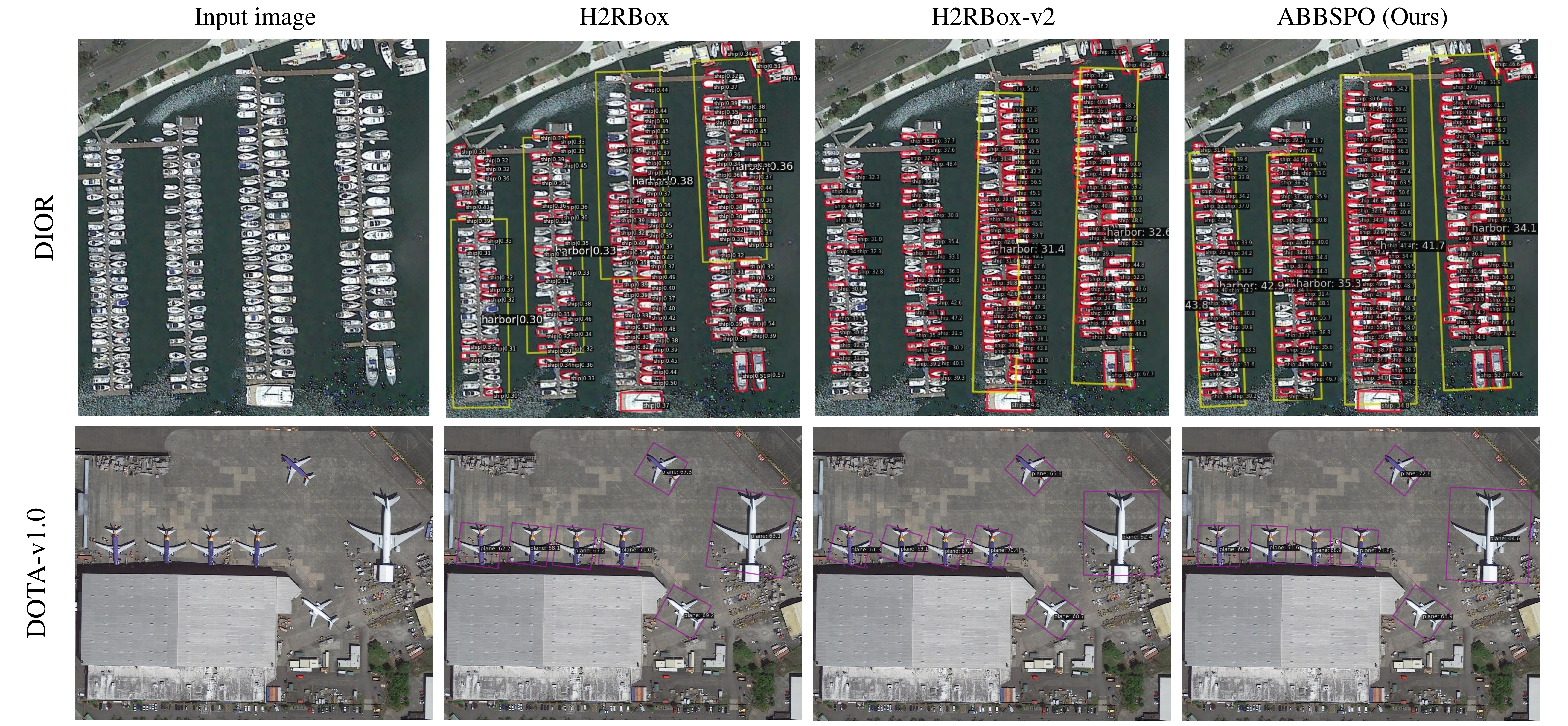}
    \caption{Qualitative OOD results on DIOR \cite{dior, dior-r} and DOTA-v1.0 \cite{dota} datasets.}
    \label{fig:figure10}
\end{figure*}
\begin{figure*} [htbp]
    \centering
    \includegraphics[width=\linewidth]{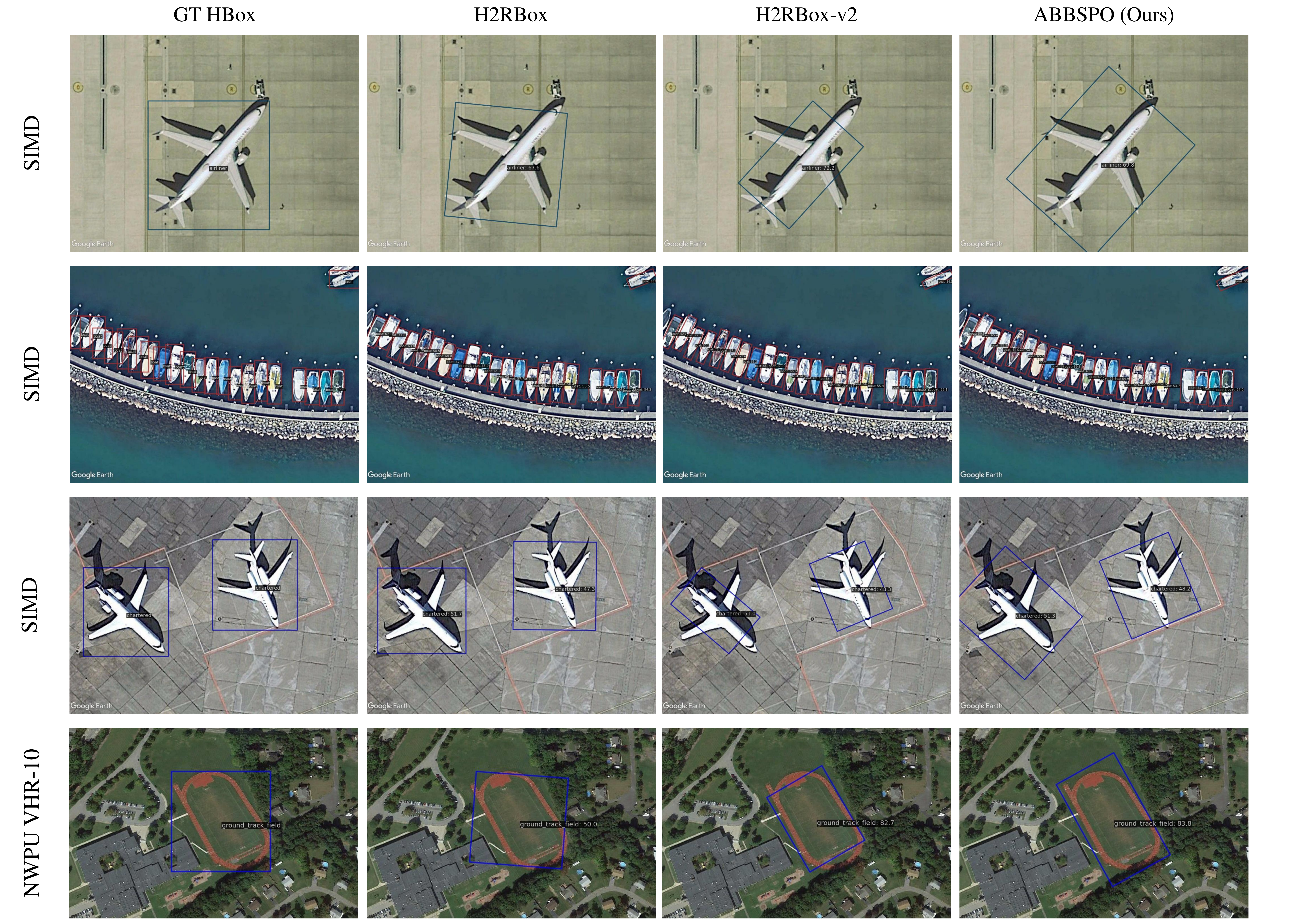}
    \caption{Qualitative OOD results on SIMD \cite{simd} and NWPU VHR-10 \cite{nwpuvhr-10} datasets.}
    \label{fig:figure11}
\end{figure*}
\clearpage

\section{Additional Qualitative Results}
\label{additional qualitative results}
\noindent\textbf{DIOR.} As shown in the first row of Fig.~\ref{fig:figure10}, our ABBSPO successfully detects more ships and harbors with higher precision. This demonstrates an improved object capturing ability of our model, achieved through an effective training process.

\noindent\textbf{DOTA-v1.0.} As shown in the second row of Fig.~\ref{fig:figure10}, our ABBSPO predicts the orientations of airplanes more accurately. This highlights an enhanced angle prediction accuracy of our ABBSPO, which is reinforced by our SPA loss-based self-supervision during training.

\noindent\textbf{SIMD \& NWPU VHR-10.} 
As shown in the first column of Fig.~\ref{fig:figure11}, the GT HBoxes for the corresponding test images are displayed since they only contain GT T-HBoxes. 
Across the four test images, our ABBSPO demonstrates superior OOD performance in terms of both scale and orientation predictions.
Notably, in the first and third rows that show predictions for airplanes, our ABBSPO successfully predicts accurate orientations and scales of the objects while H2RBox \cite{h2rbox} and H2RBox-v2 \cite{h2rbox-v2} predicts inaccurate orientations and scales for the objects, which often occurs when their models are trained with GT T-HBoxes.

\clearpage

\end{document}